\documentclass[10.5pt,peerreview,draftclsnofoot, onecolumn,compsoc]{IEEEtran}

\usepackage[OT1]{fontenc}
\usepackage{bbm}
 \usepackage{graphicx}
\usepackage{epsfig}
\usepackage{amssymb}
\usepackage{amsfonts}
\usepackage{multirow}
\usepackage{algorithm}
\usepackage{amsthm}
\usepackage{algpseudocode}
\usepackage{algorithm}
\usepackage{float}
\usepackage{caption} 
\usepackage{booktabs} 
\usepackage{mathrsfs}
\usepackage{amsmath}
\usepackage{bm}
\usepackage{bbm}
\usepackage[normalem]{ulem}
\usepackage{pdflscape}
\usepackage{graphicx}
\usepackage{adjustbox}
\usepackage{footnote}
\usepackage{color,soul}
\usepackage{tcolorbox}
\usepackage{adjustbox}

\usepackage{color, colortbl}
\definecolor{Gray}{gray}{0.97}
\usepackage{xcolor}

\usepackage{adjustbox}
\usepackage{changepage}

\usepackage[utf8x]{inputenc}

\usepackage{textcomp,marvosym}

\usepackage{cite}

\usepackage[right]{lineno}

\usepackage{array}

\usepackage[OT1]{fontenc}
\usepackage{bbm}
\usepackage{graphicx}
\usepackage{epsfig}
\usepackage{amssymb}
\usepackage{amsfonts}
\usepackage{multirow}
\usepackage{algorithm}
\usepackage{amsthm}

\usepackage{algpseudocode}
\usepackage{algorithm}
\usepackage{float}
\usepackage{caption} 
\usepackage{booktabs} 
\usepackage{mathrsfs}
\usepackage{amsmath}
\usepackage{bm}
\usepackage{bbm}
\usepackage[normalem]{ulem}
\usepackage{pdflscape}
\usepackage{graphicx}
\usepackage{adjustbox}
\usepackage{epstopdf}
\usepackage{footnote}
\usepackage{color,soul}
\usepackage{tcolorbox}
\newtcbox{\hltitle}[1][yellow]{on line, arc=7pt,colback=#1!10!yellow,colframe=#1!50!white,
 before upper={\rule[-3pt]{0pt}{10pt}},boxrule=1pt, boxsep=0pt,left=6pt,
 right=6pt,top=2pt,bottom=2pt}
\usepackage{adjustbox}

\usepackage{adjustbox}

\usepackage{natbib}

\begin{document}

\title{Bayesian Patchworks: An Approach to Case-Based Reasoning} 
%
%
%
%

\author{Ramin ~Moghaddass,~
 Cynthia~Rudin
\IEEEcompsocitemizethanks{\IEEEcompsocthanksitem R. Moghaddass is with the Department
of Industrial Engineering, University of Miami, Coral Gables, USA.\protect\\
E-mail: ramin@miami.edu
\IEEEcompsocthanksitem C. Rudin is with the Computer Science Department and Electrical and Computer Engineering Department of Duke University.
\protect\\
E-mail: cynthia@cs.duke.edu}
\thanks{}}

\IEEEtitleabstractindextext{%
\begin{abstract} 
Doctors often rely on their past experience in order to diagnose patients. For a doctor with enough experience, almost every patient would have similarities to key cases seen in the past, and each new patient could be viewed as a mixture of these key past cases. Because doctors often tend to reason this way, an efficient computationally aided diagnostic tool that thinks in the same way might be helpful in locating key past cases of interest that could assist with diagnosis. This article develops a novel mathematical model to mimic the type of logical thinking that physicians use when considering past cases. The proposed model can also provide physicians with explanations that would be similar to the way they would naturally reason about cases. The proposed method is designed to yield predictive accuracy, computational efficiency, and insight into medical data; the key element is the insight into medical data -- in some sense we are automating a complicated process that physicians might perform manually. We finally implemented the result of this work on two publicly available healthcare datasets, for (1) heart disease prediction and (2) breast cancer prediction. 
 \end{abstract}
}

\maketitle

\IEEEdisplaynontitleabstractindextext

\IEEEpeerreviewmaketitle

\IEEEraisesectionheading{\section{Introduction \& Literature Review}}
\sloppy

Doctors often rely on their past experience in order to diagnose patients, often considering similar cases from the past to diagnose a new case. A radiologist might consider how similar a medical image is to an exemplar from a medical reference book, or an image from one of their own past cases. An infectious disease specialist might consider how similar someone's symptoms are from past patients in order to diagnose an illness. For a doctor with enough experience, almost every patient would have similarities to key cases seen in the past, and each new patient could be viewed as a mixture (a patchwork) of these key past cases. Because doctors often tend to reason this way, a computationally aided diagnostic tool that thinks in the same way might be helpful in locating key past cases of interest that could assist with diagnosis.

More generally, beyond medical decisions, it is well known that various types of exemplar-based reasoning and prototyping are fundamental to the way humans make tactical decisions \citep{ carroll1980analyzing, cohen1996metarecognition, klein1989decision,newell1972human}. Humans also tend to think of categories as being represented by a specific member of that category, which is the center of the subfield of cognitive science called \textit{prototype theory} \citep{Rosch1973}. It has been shown that often, user confidence in a solution is increased when an example case is displayed rather than the logic of the decision rule alone \citep{cunningham2003evaluation}. The fact that medical students and MBAs are taught through case-based teaching methods is not a coincidence, it is because this type of thinking is how our minds work. In \textit{recognition-primed decision making} \citep{klein1989decision} for fire-fighters, students are taught to consider past cases in order to make informed decisions. By using previous experience in form of cases to understand and solve new problems, case-based reasoning has been reported as a suitable and successful technique for medical knowledge-based systems and medical decision making (see for example \cite{Schmidt2001355,Dussart2008718} and the discussion about the medical domain).


Case-Based Reasoning (CBR) (see \cite{aamodt1994case, Biswas2014235,slade1991case,Irissappane2015477}) is a subfield of artificial intelligence dedicated to automated exemplar-based decisions. CBR has been used in real-world applications for many years (e.g., \cite{Anaissi2015, Bichindaritz2006,Janssen2015,li2008ranking}), including many medical applications \cite{BegumEtAl2011,BegumEtAl09}. Researchers
 have used case-based reasoning for treatments of patients with anxiety disorders \citep{Janssen2015}, while others have used it for elder care and health assessments \citep{Hu2014}, analyzing gene expression data to study acute lymphoblastic leukemia (ALL) \citep{Anaissi2015}, and to study patient registration on renal transplant waiting lists \citep{CampilloGimenez}. 
Many case-based reasoning systems use some form of nearest neighbor method \citep{cover1967nearest} (also see \cite{Anaissi2015,cover1967nearest,Song2013}). 
Several works provide in-depth discussions of the issues and challenges of applying case-based reasoning in medical domains \citep{BegumEtAl09,BegumEtAl2011,Bichindaritz2006}.

Our approach is different than typical case-based reasoning approaches in that each new case is modeled as a mixture of different parts of past cases, where the past cases vote on the features and label of the new case. In essence, each patient is modeled as if it were a patchwork of parts of past cases. 
This approach avoids fixing a single similarity measure as in nearest neighbor approaches; we want to compare patients only on relevant subsets of features. For instance, hip-injury patients should be compared only on the similarity of their hip injuries (and age and related characteristics), and heart patients should be compared only on aspects related to heart health. The definition of a patient's neighbors should depend on how they are being compared, and different comparisons should be made based on different subsets of features.
Our method performs joint inference on the selection of past cases that are influential to each new case, the subset of features of the past cases that are relevant to the new case, and how strong the influences of the past cases are on the new case. Because we use Bayesian hierarchical modeling, the hierarchy allows us to borrow information across cases about which features are important for new cases. The hierarchical Bayesian setting helps to quantify uncertainty, and yields reasonable estimates of the (meaningful) parameter posteriors, and avoids over-fitting. 

We hope that by providing doctors with the logic they would have considered anyway, had they been able to do lengthy querying of the database for each case, doctors will be more likely to trust the model and make personalized decisions that are both evidence-based and experience-based. This method can help train doctors as well, by showing them how the key elements of past cases are relevant. 
It should be pointed out that our model can be applied in other application settings where interpretability of the model is of some importance. Examples are criminal justice, and energy reliability. We should point out that for policy-based decisions that rely on subpopulations, our model may not be the right approach.

In Section \ref{SecStruc}, the structure of the main model and the developed hierarchical framework are presented, along with a useful extension. In Section \ref{SecExperiments} we show the result of our models on two healthcare datasets which are publicly available. We also compare our model with some other important machine learning algorithms in Section \ref{SecExperiments}. 
Several appendices include details of the inference procedure.

\vspace{-.1in}
\section{Structure of Bayesian Patchworks\label{SecStruc}}

\begin{figure}[]
\centering
\includegraphics[scale=.43,trim=0cm 0.9cm 0.cm .7cm]{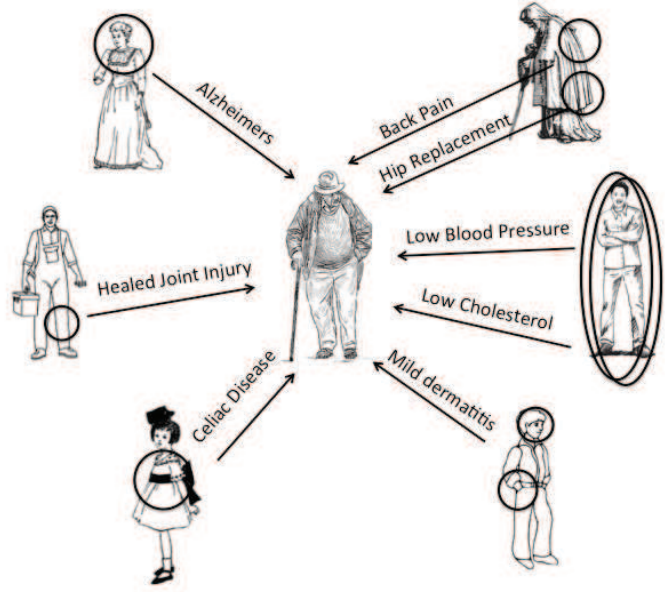}
\caption{Illustration of the Bayesian Patchworks model. Each feature of a new case (the new case is represented by the person in the center) is considered with respect to other cases with similar features (people around the outside). When we consider an aspect of the new patient's health (e.g., his skin), we would look at other patients who are similar in that aspect (for instance the patient on the lower right who has a similar case of mild dermatitis). \label{ClipArtFig}}\vspace{-.15in}
\end{figure}

Each patient (``case") is represented by a vector of features, including history of treatments, genetic makeup, environmental factors, and lifestyle.
In this model, each patient is comprised of important pieces of related individuals. These special individuals, called \textit{neighbors}, belong to a large historical set of past cases called the \textit{parent set}. Stated in terms of the generative structure of the model, each patient is generated from its neighbors, where the neighbors
give the new patient their influential features. The neighbors also vote to determine the label for the new case. Figure \ref{ClipArtFig} illustrates this, where each feature of a new patient's medical state is considered with respect to other patients (parents) that are similar in that feature.

Thinking in reverse (starting from the patient, rather than its generative process), this form of mathematical modeling naturally leads to answers to the following questions: For a new patient with unique characteristics/features, what are important similar past cases?
How close are these past cases to the current patient and what are the special factors that make this past case resemble the current case? How can this similarity help with inference -- can we infer the patient's health condition and possible treatments using the influential neighbors? The answers to these questions
are learned (inferred) from the data with our hierarchical model. We introduce the main important variables below. After that we will introduce the full-blown generative model.


Each individual has $P$ features (e.g., age, race, gender, treatment or diagnosis-related features, drugs), and we denote the $j^{th}$ feature of the $i^{th}$ observation as $x_{ij}$. The $j$th feature can take any of its $V_j$ possible discrete outcomes. The set of feature values associated with the $i$th individual are denoted by vector $\mathbf{x}_i$. 
Each individual is associated with one of the $M$ possible health classes/outcomes, which we denote for the $i$th individual by $y_i$. The whole dataset is then denoted by ($\mathbf{X}$,$\mathbf{Y}$). The main parameters of the model are as follows: 
\vspace{.1in}

\noindent$z_{ib}$: This is an indicator variable denoting whether patient $b$ from the parent set is in the neighbor set of individual $i$ (we say ``$b$ is a neighbor of $i$"). A patient and its neighbors share at least some important common properties. The neighbors for several cases can be the same. We have $z_{ib} \in\{0,1\}, i\in\{1,\cdots,N\},$ and $b\in\{1,\cdots,S\}$ and $N$ is the number of current cases, $S$ is the number of parents (past cases).

\vspace{.15in}

\noindent$w_{ibj}$: This binary parameter (where $i\in\{1,\cdots,N\}; b\in\{1,\cdots,S\}; j\in\{1,\cdots,P\}$) shows whether feature $j$ of neighbor $b$ is important to individual $i$, and thus shows whether two cases (a new case and a case from the pool of past cases in the parent set) are connected to each other. 
 
%
\vspace{.15in}

\noindent$g_{ij}(v)$: This determines the probability that feature $j$ of patient $i$ takes value $v$. Specifically, it determines how likely outcome $v$ will be copied from the neighbors. The vector $g_{ij}$ for the $v$th outcome of feature $j$ ($j \in \{1,\cdots,V_j\}$) is calculated as follows: 
\begin{equation}
\label{eq:g}
g_{ij}(v)=\lambda_0+\lambda \sum_b \mathbf{1} {[z_{ib}=1 \quad \text{and}\quad w_{ibj}=1 \quad \text{and} \quad x_{bj}=v ]}.
\end{equation}
Based on this definition, we have a boost in the score $g_{ij}(v)$ for the $v$th outcome of feature $j$ for any neighbor $b$ who has outcome $v$ for feature $j$. Here $\lambda_0$ is the baseline rate of seeing outcome $v$, and $\lambda$ determines how much influence each neighbor $b$ has. 

\vspace{.15in}

\noindent $h_i(m)$: This determines how much the neighbors of individual $i$ influence it to choose the $m$th class label ($m \in \{1,\cdots,M\}$). This variable is defined as
\begin{equation}\label{eq:h}
h_{i}(m)=\mu_0+\mu_m \sum_b \mathbf{1} {[z_{ib}=1 \quad \text{and}\quad y_{b}=m ]}.
\end{equation}
Here, $\mu_0$ and $\mu_m (m \in {1,...,M})$ are the hyperparameters that together reflect how neighbors can affect the generation of a label. In this paper we assumed that $\mu_m=\mu \;\;\forall m \in {1,...,M}$. If $\frac{\mu_0}{\mu}$ is very large, then neighbors are not important in determining the class of each individual. We will discuss extensions of this model later in the paper. 

Consider the generation process for individual $i$. After the values $\bm{z} = [z_{ij}]$ and $\bm{w}=[w_{ibj}]$ are generated for $j \in \{1,...,J\}$ and $b\in\{1,...,S\}$, we know $i$'s neighbors and their influences. From there, $\bm{g}_{ij}(:)$ generates vector $(\bm{\phi}_{ij}(:))$ and then finally $\bm{\phi}_{ij}(:)$ generates the value of the $j$th feature of individual $i$ (i.e., $x_{ij}\sim \textrm{Cat} (\bm{\phi}_{ij}(:)), \forall j \in \{1,...,J\}$). Also from $\bm{z}$, we know $\bm{h}_i(:)$ that generates the outcome for individual $i$ (i.e., $y_i\sim \textrm{Cat}(\bm{\theta}_i)$). 

As an example, we consider a patient $i$. All of the past patients with common properties are neighbors, so that $z_{ib}=1$. The first neighbor has a circulatory system condition that is similar to patient $i$. The second neighbor has a hip injury similar to patient $i$, and so on. Each of these medical conditions of the neighbors provide some information about patient $i$'s features and labels. For the information from neighbor $b$ that definitely informs feature $j$ for patient $i$, we would have $w_{ibj}=1$, and the probability that patient $i$ would have the same value $v$ of the feature as the neighbor would be large, meaning $g_{ij}$ would be large for value $v$. If neighbor $b$ influences $i$'s class label, then $h_i$ would be large for neighbor $b$'s class label. This way, we can consider a patient as a patchwork of past patients that are similar to the present patient. The neighbors' important characteristics (the hip injury, the circulatory condition, etc.) all contribute to the features and outcomes for the present patient. An understanding of these similar past patients serves as a point of reference for how to care for the present patient.

\subsection{Full Hierarchical Bayesian Patchworks Model (Model I)} \label{sec:fullhier}

The variables of the hierarchical model (herein denoted BPatch) are as follows: 

\vspace{.15in}
\noindent $z_{ib}$: Here $z_{ib}$ ($i\in \{1,\cdots,N\}, b\in \{1,\cdots,S\}$) follows a Bernoulli distribution with parameters $\alpha$, where hyperparameter $\alpha$ denotes how many neighbors each individual tends to have on average. This parameter controls how many parents generate each individual. Note that if $\alpha$ is chosen to be large, then the model may overfit by choosing many cases from the parent set as neighbors. 
\vspace{.15in}

\noindent $q_{j}$: This parameter determines the general importance of feature $j$ ranging from 0 to 1, 
$$q_{j}\sim \textrm{Beta}(\gamma,\sigma_1).$$ 
The hyperparameters $\gamma$ and $\sigma_1$ control the number of features used for generating the new observations and their class labels.

\vspace{.15in}

\noindent$\tilde{q}_{bj}$: This parameter determines the importance of feature $j$ to parent $b$. Its values follow a Beta distribution centered around mean $q_j$, that is,
 $\tilde{q}_{bj}\sim\textrm{{Beta}}(\sigma_2 \,q_j/(1-q_j),\sigma_2)$. (To see that the mean is $q_j$ note that the mean of the Beta$(\alpha,\beta)$ distribution is $1/(1+\beta/\alpha)$.)
 Here, $\sigma_2$ is another hyperparameter that governs the sparsity of the distribution $\tilde{q}_{bj}$. The intuition behind this parameter is that if it is set to a value that makes the variance of the $\textrm{{Beta}}(\sigma_2 \,q_j/(1-q_j),\sigma_2)$ too large, then some neighbors may be very influential for feature $j$ and some neighbors will not be influential. If it is set to a value that makes the variance too small, then neighbors may all be approximately equally influential for feature $j$.
 \vspace{.15in}

\noindent $w_{ibj}$: As explained earlier, this binary parameter shows whether feature $j$ of neighbor $b$ is important to individual $i$. This is determined via:
\[w_{ibj}\sim \textrm{Bernoulli}(\tilde{q}_{bj}).\]
\noindent This allows for individual level effects on neighbor $b$'s influence for feature $j$.

\vspace{.15in}

\noindent $\phi_{ij}$: 
The distribution of values for feature $j$ for individual $i$ comes directly from the $g_{ij}$'s (as given in Eq. \eqref{eq:g}), that is 
$$\bm{\phi}_{ij}\sim \text{Dirichlet}(g_{ij}(:)),$$ which means 
$ {\phi}_{ij}(v) \geq 0, \forall (i,j,v)$ and $\sum \limits_v \phi_{ij}(v)=1$. So, $\bm\phi_{ij}$ is the distribution of feature values for the $j$th feature of the $i$th individual. 

\vspace{.15in}

\noindent$x_{ij}$: The $j$th feature of the $i$th individual, $x_{ij}$, is generated from a categorical distribution with parameter vector $\bm{\phi}_{ij}(:)$. Therefore, the probability that $x_{ij}$ takes its $v$th outcome becomes $\phi_{ij}(v)$. 
\vspace{.15in}

Now that we have generated the observations, we need only to generate the class labels. 

\vspace{.15in}

\noindent $\bm{\theta}_i(m)$: The distribution of class outcomes for individual $i$ will be determined by $h_i$ defined in Eq. (\ref{eq:h}) through
$$\bm{\theta}_i(:)\sim \text{Dirichlet}(h_i(:)).$$ 
This way, $\theta_i(m)\geq 0$ for all $(i,m)$ and $\sum \limits_m \theta_i(m)=1$ for all $i$. 

\noindent$y_{i}$: The distribution of class outcomes for individual $i$ is generated from a categorical distribution with parameter vector $\bm{\theta}_i$. To use our model for classification, 
the class with the highest probability $\theta_i(m)$ can be used as the predicted class. 

\vspace{.15in}

\begin{figure}[]
\centering
  \includegraphics[scale=.35,trim=1cm 0cm 1.cm 0cm]{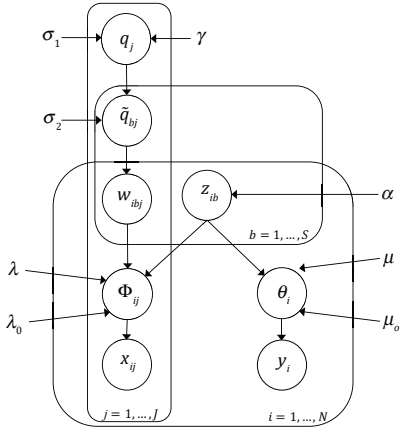}
 \caption{Plate diagram (directed acyclic graph) for an influential neighbor model (Model I). Circles indicate stochastic nodes; squares indicate fixed nodes (hyperparameters); directed edges (arrows) indicate the direction of influence. In this structure $z_{ib}$, $b=\{1,\cdots,N\}$ is one of the two key indicator variables denoting if neighbor $b$ from the past cases is an influential neighbor of individual $i$. The 2nd key variable is $w_{ibj}$, which shows the importance of feature $j$ of neighbor $b$ on affecting individual $i$. The outcomes of this generative model are the $x_{ij}$'s and $y_i$'s.}
\label{fig:prop1}\vspace{-.1in}
\end{figure}

According to this model, neighbors of an individual $i$ determine both its feature values $\mathbf{x}_i$ and its outcome $y_i$. The hyperparameters of our model are $(\alpha, \gamma, \sigma_1, \sigma_2, \lambda_0, \lambda, \mu_0, \mu)$. The graphical model is illustrated in Fig. \ref{fig:prop1}.

 The full model given all hyperparameters is summarized below: \begin{align*} \nonumber
&z_{ib}\sim \text{Bernoulli}(\alpha) \,\, \forall (i,b), & & \\ \nonumber
&q_j \sim \text{Beta}(\gamma,\sigma_1) \,\, \forall j, \quad \tilde{q}_{bj} \sim \text{Beta}(\sigma_2 \,q_j/(1-q_j),\sigma_2)\,\, \forall (b,j), \quad w_{ibj}\sim \text{Bernoulli}(\tilde{q}_{bj}) \,\, \forall (i,b,j),\\\nonumber
& g_{ij}(v)=\lambda_0+\lambda \sum_b \mathbf{1} {[z_{ib}=1 \quad \text{and}\quad w_{ibj}=1 \quad \text{and} \quad x_{bj}=v ]}, \\\nonumber & \bm{\phi}_{ij}(:)\sim \text{Dirichlet}(\bm{g}_{ij}(:)) \,\, \forall (i,j), \quad \quad x_{ij}\sim \textrm{Cat} (\bm{\phi}_{ij}(:)) \,\, \forall (i,j), \\ \nonumber
& h_{i}(m)=\mu_0+\mu \sum_b \mathbf{1} {[z_{ib}=1 \quad \text{and}\quad y_{b}=m ]}, 
\\ \nonumber
& \bm{\theta}_i\sim \text{Dirichlet}(\bm{h}_i(:)) \,\, \forall i, \quad \quad y_i\sim \textrm{Cat}(\bm{\theta}_i) \,\,\forall i.
\end{align*}

To summarize, the $z_{ib}$ determine who the relevant neighbors are for $i$, the next three lines determine how the features are generated from the values of the neighbors, and the two lines after that determine how the labels are generated. Inference for the model is discussed in Appendix \ref{SecInference}.

\subsubsection{{Continuous Features}}

Model I can be edited to handle continuous features in several ways. 
One option is to use a bandwidth or a binary distance function that determines whether two feature values in the continuous domain are close:
\begin{equation*}\label{eq:g2}
g_{ij}(v)=\lambda_0+\lambda \sum_b \mathbf{1} {[z_{ib}=1 \quad \text{and}\quad w_{ibj}=1 \quad \text{and}\quad \mathbf{I}_{j}(x_{bj},v)] }, 
\end{equation*}
where $\mathbf{I}_{j}$ is a binary similarity measure/kernel for the $j$th feature.
A simple example of such measure is a binary kernel with a bandwidth $r_j$ as follows: 
\begin{equation*}
\mathbf{I}_{j}(x_{bj},v)= \propto \left\{ \begin{array}{ll}
 1, & \mbox{if $|x_{bj}-v|\leq r_j$} \\
 0, & \mbox{otherwise}
.\end{array} \right.
\end{equation*}
From there, $\phi_{ij}$ and $x_{ij}$ values could be generated from a normal distribution or another continuous distribution.

\subsubsection{More Complicated Outcome Relationships}
Once the features $x_{ij}$ are generated, if desired one could replace the outcome model with any other standard generative modeling procedure (e.g., linear regression model or logistic regression). We chose a simple voting model for the outcome, where each neighbor casts one vote for their own outcome.

\subsubsection{{Handling Missing Points}}
 
Handling missing data is not the focus of the paper, one may handle it in the usual way, for instance, by performing any method for imputation. Other than the above method, our model can be updates as follows to deal with missing points: for each dimension of features, we can add an artificial output representing missing value. For example for feature $j$ with $V_j$ possible discrete outcomes, we add an artificial outcome $V_j+1$ and replace all missing values in that direction with it.

\subsection{Comparison with closely related work}

Now that we have introduced the full model, we can give a more technical description of how the models presented above differ from ordinary nearest neighbor techniques (k-nearest neighbor -- KNN) as well as the Bayesian Case Model (BCM) \citep{KimRuSh14}, and Latent Dirichlet Allocation (LDA).

\textbf{Nearest Neighbors:} In KNN, adaptive nearest neighbors, or nearest prototype models, there is a single fixed distance metric used for comparing a current observation to past observations. Here, the choice of distance metric depends on the observation. This allows, for instance, a patient with a heart condition and sports injury to be neighbors of patients with a similar heart condition (regardless of sports injury), and patients with a sports injury (regardless of heart condition). Our method can be viewed as a version of nearest neighbors that is much more flexible, in terms of how many neighbors are used, the distance metric between neighbors, and the completeness of the characterization of the new patient by the neighbors.



\textbf{Bayesian Case Model:} First, BCM is unsupervised, while BPatch is supervised. More importantly, BPatch includes \textit{individual level} effects for influences of parents, whereas BCM uses prototypes instead. 
 BCM limits new cases to be compared with only certain learned past cases (the prototypes). An easy way to see this is that BPatch has variables $w_{ibj}$ for each unit, controlling whether feature $j$ of neighbor $b$ is important for observation $i$. In BCM, the analogous parameter $w_{sj}$ only determines whether feature $j$ is important for cluster $s$, and there is no individual level effect. As a result, the whole model for BPatch is different: in order to determine the distribution of values for any feature $j$ (which is $\phi_{ij}$), BPatch needs to consider how many parents are going to generate it (controlled by $z_{ib}$). In contrast, BCM's $\phi_s$ values depend only on the cluster $s$. For BCM, $z_{ij}$ takes on values that are single cluster indices (``Which {cluster} will generate feature $j$ for individual $i$?"), whereas for BPatch, $z_{ib}$ determines which \textit{parents} generate individual $i$ (``Which parents will be allowed to generate individual $i$'s features?"). As a result, BPatch's generation of features is a result of a vote among parents, whereas BCM's generation of features is a result of a single prototype sharing its feature with the new observation.
 By considering individual level effects, generated for each individual as a patchwork of its parents, BPatch is conceptually different than BCM, which is cluster-based. BPatch is more similar to nearest neighbors than prototype- or cluster-based modeling like BCM.

Note that there are papers dealing with supervised prototype classification other than BCM (e.g., \cite{bien2011prototype}), though not using the important parts of each case like BCM and BPatch; this is problematic, and causes these algorithms to choose a huge number of prototypes in practice. It also limits us to compare with a single nearest prototype even if our current observation is clearly a mixture of different past situations.

\textbf{Latent Dirichlet Allocation:} LDA has a strong (and false) assumption that each feature value is generated independently. BPatch and BCM do not have this assumption, and the generation of features is correlated (either through clusters or parents). BCM and BPatch are more complicated and more realistic models, leading to better performance. This was discussed via experiments by Kim et al. \citep{KimRuSh14}.



\subsection{Model II}
\begin{figure}[]
\centering
 \includegraphics[scale=.4, trim=1cm 0cm 1.cm 0cm]{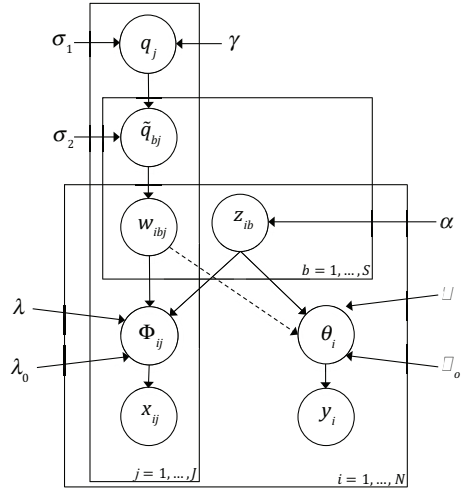}
 \caption{Plate diagram for Model II. Circles indicate stochastic nodes; squares indicate fixed nodes; directed edges (arrows) indicate parent–child relations.}
 \label{fig:ext1}\vspace{-.1in}
\end{figure}

When generating feature values or outcomes, the previous BPatch model (Model I) counts each vote from a neighbor equally, whereas in Model II, the votes are weighted by the \textit{degree of influence} of the neighbor. For instance, if one neighbor is influential for several features of the current patient, then it should potentially have a stronger vote than other neighbors in determining the feature values and outcomes.
The degree of influence of a neighbor depends on the influences of its individual features. We denote $\kappa_{ib},0 \leq \kappa_{ib} \leq P$, as the degree of influence between case $i$ and parent $b$, which is 
\[\kappa_{ib}= z_{ib} \sum_j w_{ibj}.\]
The higher $\kappa_{ib}$ is for parent $b$, the stronger its influence is on individual $i$. This is different from Model I where the influence of parent $b$ on individual $i$ was determined through a binary indicator $z_{ib}$. In Model II, the amount that each influential neighbor can potentially affect feature $j$ of individual $i$ is defined as
\begin{equation}\label{eq:gnew}
g^{new}_{ij}(v)=\lambda_0+\lambda \sum_b \kappa_{ib} \, \mathbf{1}\{w_{ibj}=1 \, \text{and} \, x_{bj}=v \}, 	
\end{equation}
and similarly the vote for the outcome can be weighted as 
\begin{equation}\label{eq:hnew}
h^{new}_{i}(m)=\mu_0+\mu \sum_b \kappa_{ib} \mathbf{1}\{y_{b}=m \}.
\end{equation}
Based on this model, not only how many neighbors vote for a feature matters but also how strongly they vote matters too. The rest of the parameters are the same as the model given in Section 2. In this model, the degree of influence directly depends on how similar each pair of individuals are. Based on this definition, if $z_{ib}=0$, then $\kappa_{ib}=0$, and if $z_{ib}\neq 0$ then $\kappa_{ib}$ equals $\sum \limits_j w_{ibj}$. The directed acyclic graph of this model is given in Fig. \ref{fig:ext1}.

The variable $\bm{\kappa}$ in this model is not a parameter to be estimated as it can be calculated from $\bm{w}$ and $\bm{z}$. However the structure of the posterior and conditionals are different from the original model in the sense that $y_i$ now depends also on all $w_{i::}$. 

The posterior for Model II is derived in Appendix \ref{app:ModelIIPosterior}.

We also propose a possible third model (Model III) in Appendix \ref{app:modelIII}, where the relationship between parent $b$ and individual $i$ can have a degree/weight between 0 and 1 following a Beta distribution. This is different from both Models I and II where $z_{ib}$ was a binary variable. Model III did not tend to perform well, and we chose to omit it in our experiments. It is possible that improvements to the model might lead to better performance, though we leave that for future work.

\subsection{Health Outcome/Label Prediction}\label{sec:healthoutcome}
Let us assume that a new patient $r$ with a feature set $\bm{x}_r$ is available, for which we may or may not have the health label ${y}_r$. We can use our model to make three types of predictions for this new individual as follows:
\begin{enumerate}
\item[(I)] For a known ${\bm{x}_r}$ and unknown ${y}_r$, what are the set of influential neighbors ${z}_{rb}$, for $b \in \{1,\cdots,S\}$, and their influences $w_{rbj}$, for $j \in \{1,\cdots,V_j\}$, $b \in \{1,\cdots,S\}$?
\item[(II)] For the same setting as (I) above, what is the estimated probability of being at each health outcome, and what is the predicted outcome $\hat{y}_r$?
\item[(III)] For a set of known $({\bm{x}_r},y_r)$, what are the set of influential neighbors ${z}_{rb},\; b \in \{1,\cdots,S\}$ and their influences $w_{ibj}$, for $j \in \{1,\cdots,V_j\},\; b \in \{1,\cdots,S\}$?
\end{enumerate}
The answers to prediction problems (I) and (II) can be obtained in a completely unsupervised way. That is, they can be answered having removed the right part of the model in Fig. \ref{fig:prop1} that generates labels $\textbf{Y}$. In other words, we do not need the outcomes of the parent set for inference in the unsupervised case. 
\section{Numerical Experiments: Applications to Healthcare Analytics}\label{SecExperiments}
We applied our models to two healthcare problems, heart disease and breast cancer predictions. Then, in Section \ref{sec:large}, we applied our models to a larger data set with more features. 



\textbf{Heart Disease Diagnosis:} 
The Cleveland database (Cleveland Clinic Foundation), obtained via \citep{Bache+Lichman:2013}, reports the presence of heart disease in 303 patients, 274 of which had no missing information, each labeled with whether they had heart disease,
and features are described in Table \ref{tab:heart}. 

\begin{table}[ht]
\centering
\caption{Description of the heart disease data used in this study
\label{tab:heart}}
\resizebox{\textwidth}{!}{%
\begin{tabular}{@{}clll@{}}
\toprule
\# & \multicolumn{1}{l}{Feature Name} & \multicolumn{1}{c}{Number of Possible Outcomes} & \multicolumn{1}{c}{Description} \\ \midrule
1 & Age & 3 & (0,45),[45,55),[55,100) \\
2 & Gender & 2 & Male, Female \\
3 & Chest Pain Type & 2 & 1 or 2, 3, 4 \\ 
4 & Trestbps (blood pressure) & 3 & (0,120], (120,140],(140, 300) \\
5 & Serum Cholestrol (Chol)& 3 & (0,240], (240,260], (260,400) \\
6 & Fasting Blood Sugar & 2 & 0, $>$0 \\ 
7 & Resting Electrographic Results (Restecg) & 2 & 0, $>$0 \\
8 & Maximum Heart Rate Achieved (Thalach) & 3 & (0,50],(50,65],(65,202)\\
9 & Exercise induced Angina (Exang) & 2 & 0, $>$0 
\\
10 & Oldpeak & 2 & (0,0.5], (.5,6.2) 
\\ 11 & Slope of the peak & 2 & 1, 2 or 3
 \\ 12 & Number of major vessels colored by fluoroscopy (Ca) & 2 & 0, 1 or 2 or 3 
 \\ 13 & Thal & 2 & 3, $>$3 \\ 
Label & Heart Disease & 2 & Presence, Absence \\
\bottomrule
\end{tabular}}
\end{table}

\textbf{Breast Cancer Recurrence Prediction:}
We used a dataset provided by the Oncology Institute, University Medical Centre, Ljubljana, Yugoslavia \citep{Bache+Lichman:2013}. 
The attributes are described in Table \ref{tab:breast}, and each observation was labeled as to whether it led to recurrence of breast cancer. After removing patients with missing values, we randomly selected a balanced set of 162 patients (81 from each class) for our analysis. 

\begin{table}[ht]
\centering
\caption{Description of the breast cancer data used in this study
\label{tab:breast}}\resizebox{\textwidth}{!}{
\begin{tabular}{@{}clll@{}}
\toprule
\# & \multicolumn{1}{c}{Feature Name} & \multicolumn{1}{c}{Number of Outcomes} & \multicolumn{1}{c}{Description} \\ \midrule
1 & Age & 3 & (0,50),[50,59),[60,100) \\
2 & Menopause & 3& $<$40, $\geq$40, premeno \\
3 & Tumor-size & 4 & [0,19], [20,29],[30,39], [40,59] \\ 
4 & Inv-nodes (blood pressure) & 2 & (0,2], $>$2 \\
5 & Node-Caps & 2 & yes, no \\
6 & Deg-malig & 3 & 1,2,3 \\
7 & Breast & 2 & left, right \\
8 & Breast-Quad & 4 & left-up, left-low, right-up, or right-low, central\\
9 & Irradiant & 2 & yes, no \\
 Label & class distribution & 2 & no-recurrence-events, recurrence-events \\
\bottomrule
\end{tabular}}
\end{table}

 \subsection{{Prediction Results}}

Both supervised and unsupervised versions of BPatch perform about equally well in predicting patients with cancer and heart disease. This indicates that learning a good set of neighbors is useful not only for characterizing the current patient's features, but also for predicting their label. Let us show this.

The hyperparameters were chosen as follows: 
$\alpha=0.5$, $\sigma_1=5$, $\gamma_1=0.5$, $\gamma_2=0.5$, $\mu_0=0.001$, $\mu=1$, $\lambda_0=0.001$, $\lambda=2$. 
The number of parents was initially chosen to be 80 ($S=80$) for both datasets. Parents were randomly selected from the training sets in each fold. In Section \ref{sec:sens}, we discuss the sensitivity of the prediction results with respect to all hyperparameters and discussed why these parameters are interpretable in that they can control both the structure of the model and its accuracy/prediction power. Prediction evaluation was performed in two ways, unsupervised and supervised. For the unsupervised analysis, the model structure was trained without using the outcome of any of the patients in the training set and only the vector of patient attributes was used to infer model parameters. In other words, the neighbors and their influences were found solely from the feature values (see Section \ref{sec:healthoutcome}). Note that while the training process itself is unsupervised, we did use the trained parameters along with the labels of training points in order to make predictions for the cases in the test set. The training set determines how we should combine the neighbors' labels to produce a label for the current case. In the supervised case, we used the information on the actual health outcomes of the patients in the training set both to train the model and to make predictions for the new cases. For both cases, we performed 5-fold cross-validation; here we divided our data into five folds, each with the same number of parents, trained our models on four folds and tested on the fifth.\footnote{The individuals in the parent list were chosen uniformly at random in each fold. We repeated our experiments with fixed sets of parents over all folds. The results were similar to models trained with different parents in each fold (omitted here).} We performed all experiments for both data sets using Model I and Model II. After training, we used the posterior predictive distribution of $\bm{\theta}$ to estimate the probability of having heart disease and breast cancer (as a recurrent event) for all patients in the test set. 
 Using the estimated probability $\bm{\theta}_i$ for patient $i$ as a classification input (where 0.5 is the threshold), we then calculated the classification accuracy, sensitivity, specificity, precision, recall, and F-measure (see Tables \ref{tab:result1} and \ref{tab:result2}). We reported these measures for each fold and followed by the mean and standard devision over folds. 


\begin{table}[]
\centering
\caption{Prediction results for the heart disease dataset using Models I and Model II}\label{tab:result1}
\resizebox{\textwidth}{!}{%
\begin{tabular}{@{}c|cc|cc|cc|cc|cc@{}}
\toprule
\multicolumn{11}{c}{\textbf{Supervised}} \\ \midrule
\multicolumn{1}{r}{Measure} & \multicolumn{2}{c}{Accuracy} & \multicolumn{2}{c}{Sensitivity} & \multicolumn{2}{c}{Specificity} & \multicolumn{2}{c}{Precision} & \multicolumn{2}{c}{Recall} \\ \midrule
\multicolumn{1}{l}{Fold} & Model I & Model II & Model I & Model II & Model I & Model II & Model I & Model II & Model I & Model II \\ \midrule
1 & 74.55 & 81.82 & 73.33 & 73.33 & 76.00 & 92.00 & 78.57 & 91.67 & 73.33 & 73.33\% \\
2 & 78.18 & 76.36 & 80.00 & 80.00 & 76.00 & 72.00 & 80.00 & 77.42 & 80.00 & 80.00\% \\
3 & 78.18 & 78.18 & 69.23 & 73.08 & 86.21 & 82.76 & 81.82 & 79.17 & 69.23 & 73.08\% \\
4 & 92.73 & 92.73 & 92.31 & 92.31 & 93.10 & 93.10 & 92.31 & 92.31 & 92.31 & 92.31\% \\
5 & 90.74 & 88.89 & 96.00 & 100.00 & 86.21 & 79.31 & 85.71 & 80.65 & 96.00 & 100.00\% \\ \midrule
\rowcolor{Gray}
Mean & 82.88 & \textbf{83.60} & 82.17 & \textbf{83.74} & 83.50 & \textbf{83.83} & 83.68 & \textbf{84.24} & 82.17 & \textbf{83.74}\% \\
SD & 8.25 & 7.01 & 11.66 & 11.98 & 7.41 & 8.86 & 5.52 & 7.17 & 11.66 & 11.98\% \\ \bottomrule
\toprule
 \multicolumn{11}{c}{\textbf{Unsupervised}} \\ \midrule
\multicolumn{1}{r}{Measure} & \multicolumn{2}{c}{Accuracy} & \multicolumn{2}{c}{Sensitivity} & \multicolumn{2}{c}{Specificity} & \multicolumn{2}{c}{Precision} & \multicolumn{2}{c}{Recall} \\ \midrule
\multicolumn{1}{l}{Fold} & Model I & Model II & Model I & Model II & Model I & Model II & Model I & Model II & Model I & Model II \\ \midrule
1 & 72.73 & 72.73 & 73.33 & 73.33\% & 72.00 & 72.00 & 75.86 & 75.86 & 73.33 & 73.33\% \\ 2 &76.36 & 74.55 & 76.67 & 76.67\% & 76.00 & 72.00 & 79.31 & 76.67 & 76.67 & 76.67\% \\ 3 & 80.00 & 80.00 & 73.08 & 76.92\% & 86.21 & 82.76 & 82.61 & 80.00 & 73.08 & 76.92\% \\ 4 & 89.09 & 94.55 & 84.62 & 100.00 & 93.10 & 89.66 & 91.67 & 89.66 & 84.62 & 100.00\%
\\
5 & 87.04 & 87.04 & 92.00 & 92.00\% & 82.76 & 82.76 & 82.14 & 82.14 & 92.00 & 92.00\% \\ \midrule
\rowcolor{Gray}
Mean & 81.04 & \textbf{81.77} & 79.94 & \textbf{83.78} & \textbf{82.01} & 79.83 & \textbf{82.32} & 80.87 & 79.94 & \textbf{83.78}\% \\
SD & 6.94 & 9.06 & 8.20 & 11.59 & 8.33 & 7.69 & 5.88 & 5.53 & 8.20 & 11.59\% \\ \bottomrule
\end{tabular}}
\end{table}

\begin{table}[]
\centering
\caption{Prediction results for the breast cancer dataset using Models I and II}
\resizebox{\textwidth}{!}{%
\begin{tabular}{@{}c|cc|cc|cc|cc|cc@{}}
\toprule
\multicolumn{11}{c}{\textbf{Supervised}} \\ \midrule
\multicolumn{1}{r}{Measure} & \multicolumn{2}{c}{Accuracy} & \multicolumn{2}{c}{Sensitivity} & \multicolumn{2}{c}{Specificity} & \multicolumn{2}{c}{Precision} & \multicolumn{2}{c}{Recall} \\ \midrule
\multicolumn{1}{l}{Fold} & Model I & Model II & Model I & Model II & Model I & Model II & Model I & Model II & Model I & Model II \\ \midrule
1 & 68.75 & 71.88 & 92.86 & 92.86 & 50.00 & 55.56 & 59.09 & 61.90 & 92.86 & 92.86\% \\
2 & 90.91 & 87.88 & 100.00 & 100.00 & 80.00 & 73.33 & 85.71 & 81.82 & 100.00 & 100.00\% \\
3 & 62.50 & 71.88 & 57.14 & 71.43 & 66.67 & 72.22 & 57.14 & 66.67 & 57.14 & 71.43\% \\
4 & 84.38 & 81.25 & 87.50 & 81.25 & 81.25 & 81.25 & 82.35 & 81.25 & 87.50 & 81.25\% \\
5 & 84.85 & 81.82 & 94.74 & 89.47 & 71.43 & 71.43 & 81.82 & 80.95 & 94.74 & 89.47\% \\ \midrule
\rowcolor{Gray}
Mean & 78.28 & \textbf{78.94} & 86.45 & \textbf{87.00} & 69.87 & \textbf{70.76} & 73.22 & \textbf{74.52} & 86.45 & \textbf{87.00}\% \\
SD & 12.04 & 6.95 & 16.98 & 11.01 & 12.65 & 9.36 & 13.89 & 9.50 & 16.98 & 11.01\% \\ \bottomrule
\toprule
\multicolumn{11}{c}{\textbf{Unsupervised}} \\ \midrule
\multicolumn{1}{r}{Measure} & \multicolumn{2}{c}{Accuracy} & \multicolumn{2}{c}{Sensitivity} & \multicolumn{2}{c}{Specificity} & \multicolumn{2}{c}{Precision} & \multicolumn{2}{c}{Recall} \\ \midrule
\multicolumn{1}{l}{Fold} & Model I & Model II & Model I & Model II & Model I & Model II & Model I & Model II & Model I & Model II \\ \midrule
1 & 76.36 & 71.88 & 76.67 & 92.86 & 76.00 & 55.56 & 79.31 & 61.90 & 76.67 & 92.86\% \\
2 & 76.36 & 87.88 & 76.67 & 100.00 & 76.00 & 73.33 & 79.31 & 81.82 & 76.67 & 100.00\% \\
3 & 59.38 & 65.63 & 57.14 & 71.43 & 61.11 & 61.11 & 53.33 & 58.82 & 57.14 & 71.43\% \\
4 & 89.09 & 84.38 & 84.62 & 87.50 & 93.10 & 81.25 & 91.67 & 82.35 & 84.62 & 87.50\% \\
5 & 78.79 & 84.85 & 89.47 & 94.74 & 64.29 & 71.43 & 77.27 & 81.82 & 89.47 & 94.74\% \\ \midrule
\rowcolor{Gray}
Mean & 76.00 & \textbf{78.92} & 76.91 & \textbf{89.30} & \textbf{74.10} & 68.54 & \textbf{76.18} & 73.34 & 76.91 & \textbf{89.30}\% \\
SD & 10.67 & 9.64 & 12.33 & 10.95 & 12.58 & 10.21 & 13.99 & 11.90 & 12.33 & 10.95\% 
\\ \bottomrule
\end{tabular}}
\label{tab:result2}
\end{table}

We can observe from the results in Tables \ref{tab:result1}-\ref{tab:result2} that both supervised and unsupervised version of Model I and Model II perform well in terms of classifying patients with heart disease and breast cancer conditions. The supervised version performs slightly better in both data sets, which is not surprising because it uses more information (the labels $y$ of past cases) for estimation of parameters. In the unsupervised case, recall that the labels of past cases were not used to train the model parameters, however, they were used to make predictions from the trained parameters.
That is, \textit{without using the health outcomes of past cases, our models inferred which neighbors to consider for predicting health outcomes simply by looking at neighbors with similar patterns in feature values}. 

Model II tended to have better performance than Model I in both data sets as it gives more weights to neighbors with more similar features. As discussed before, Model II is computationally more expensive than Model I and thus takes longer to converge. See Section \ref{sec:large} for the discussion on CPU analysis of two models.

\begin{figure}[]
\centering
 \includegraphics[scale=.45, trim=0cm 0cm 0.cm 0cm]{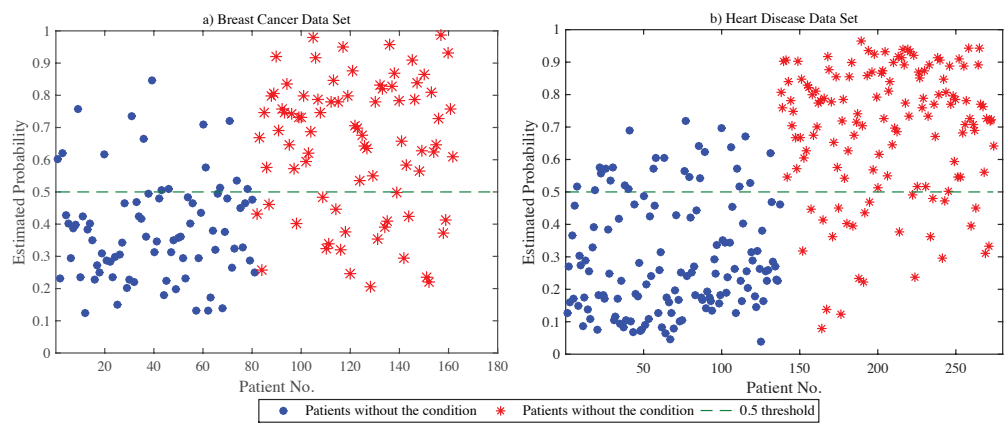}
 \caption{Estimated probability of Heart Disease and Breast Cancer Using Model II - Results are based on the cases in the 5 test sets; the dashed horizontal line is the 50\% threshold. The points on the left side of each plot (dot circles) are associated with patients with no health condition.}
\label{fig:estpr}
\end{figure}

BPatch produces probabilistic estimates for each class, unless other nearest neighbor models (and unlike BCM which is unsupervised). We plotted the estimated probability of having heart disease and a recurrent breast cancer event for the patients in the test sets in Figure \ref{fig:estpr}.
This figure shows that the probability of the health condition of interest tends to be higher for patients who actually have that condition. 



\subsection{{Discussion on Interpretability}}

Interpretability is known to be a major concern in machine learning for high-stakes decisions, but it is context-dependent, subjective, and hard to quantify. 
In what follows, we discuss three aspects of interpretability: the view of each patient through the lens of the neighbors, discussed in Section \ref{subsec:similarity}, and feature importance in Section \ref{subsec:featureimp}.
We also further provide an interpretability comparison in Section \ref{subsec:comp} to BCM, KNN, and decision trees.

\subsubsection{Insight About Cases from Bayesian Patchworks} 
\label{subsec:similarity}

\begin{figure}[]
\centering
\includegraphics[trim={6cm 0cm 0cm .1cm},scale=.5]{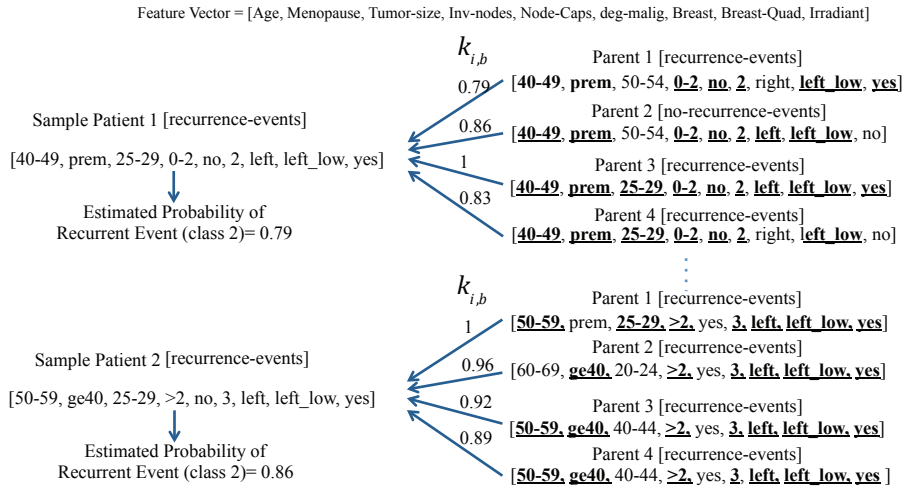}
 \caption{Example of two patients with breast cancer and 4 similar past cases in the Parents set. Note that the top four parents are shown only. The feature names are shown inside the bracket above the figure.} \label{fig:network1}
\end{figure}

\begin{figure}[]
\centering
\includegraphics[trim={6cm 0cm 0cm .1cm},scale=.5]{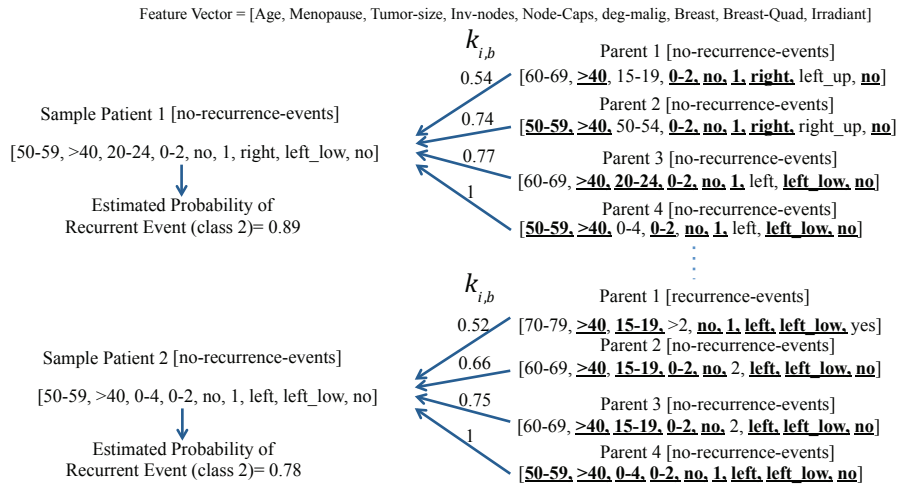}
 \caption{Example of two patients without breast cancer and four similar past cases in the Parents set.} \label{fig:network11}
\end{figure}

\begin{figure}[]
\centering
\includegraphics[trim={6cm 0cm 0cm .2cm},scale=.5]{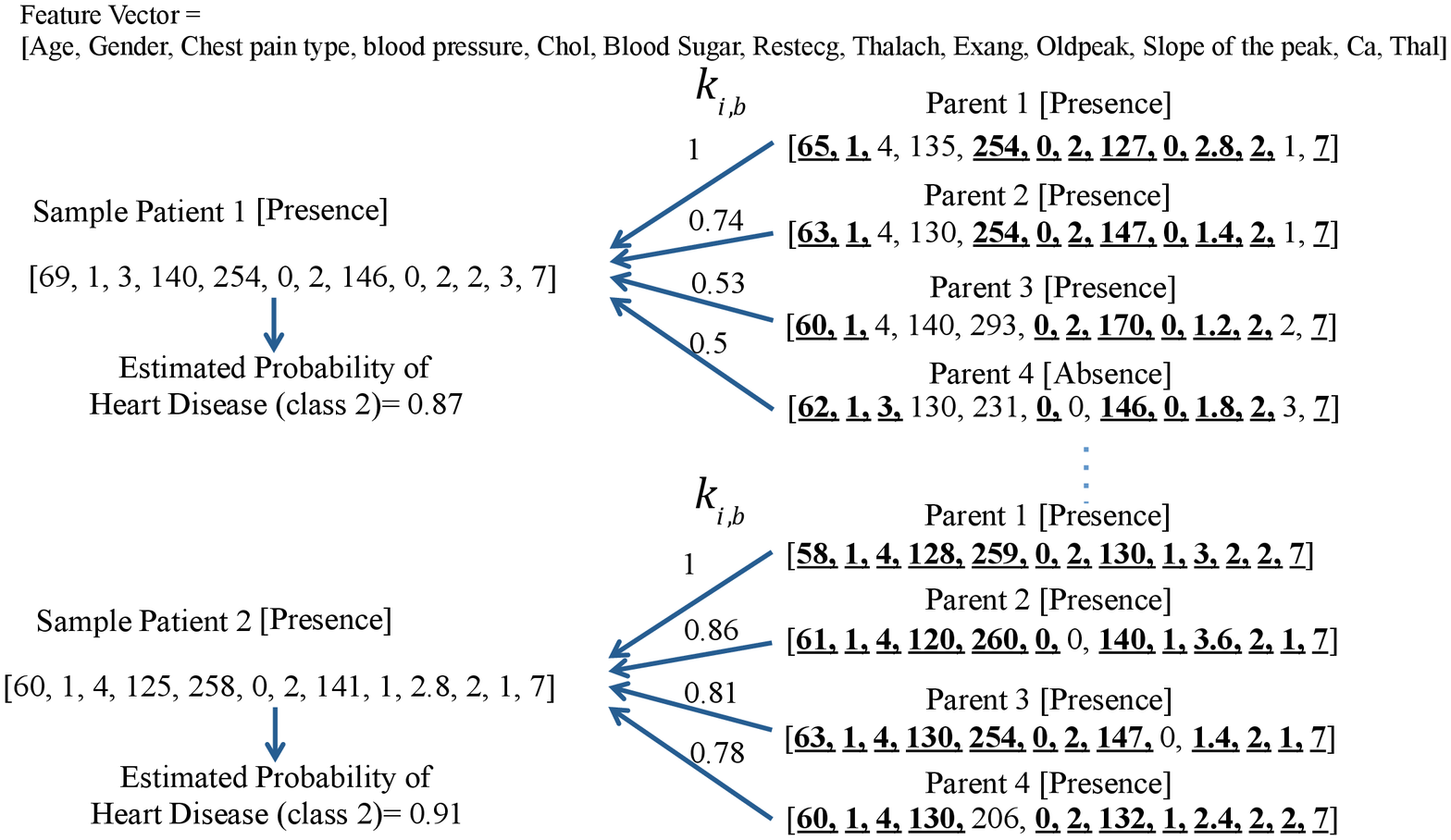}
 \caption{Example of two patients with heart disease and four similar past cases in the Parents set. The feature names are shown inside the bracket above the figure.} \label{fig:network111}
\end{figure}

\begin{figure}[]
\centering
\includegraphics[trim={6cm 0cm 0cm .1cm},scale=.5]{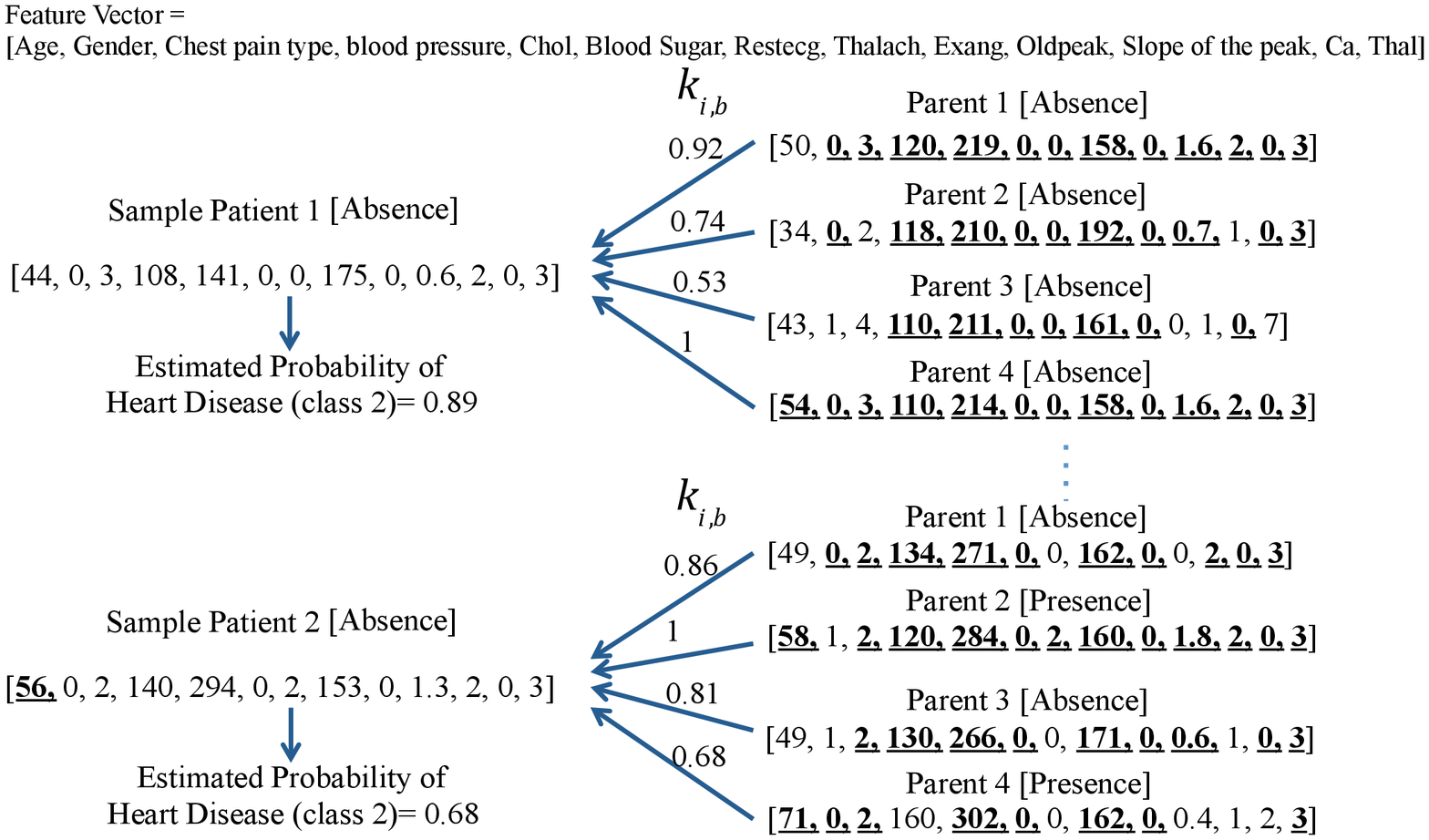}
 \caption{Example of two patients without heart disease and four similar past cases in the Parents set. The feature names are shown inside the bracket above the figure.} \label{fig:network1111}
\end{figure}

 BPatch can potentially pinpoint which past cases are similar to a new case, why certain past patients are similar to a new case, how similar they are, and how this similarity is quantified. Let us show by few examples. In Figures \ref{fig:network1}-\ref{fig:network1111}, the top neighbors/parents with the highest $\kappa_{i,b}$ are shown for four sample patients with and without breast cancer and heart disease. For each sample, the feature values of the parent associated with important dimensions are shown in bold (i.e., when $w_{ibj}=1$). It can also be seen that each parent has a unique combination of features that are important for that parent.
 The parents that are selected as neighbors tend to be similar to the patients on the important features for each parent. Consider Sample Patient 1 in Fig. \ref{fig:network1} for instance. It can be seen that its Parents 1-4 have some features identical to those of Sample Patient 1. Patients with cancer are connected more often to parents with cancer. For example, three out of the selected four parents have the same health label as Patient 1. Doctors now can look at all selected parents and see why these parents are selected as the influential neighbors and use the predicted probability of the health condition from the neighbors to make their medical decision or diagnosis. Doctor can also use these results to investigate similar past cases in terms of possible treatments in order to make better treatment decisions.

\subsubsection{Feature Importance}\label{subsec:featureimp}

Bayesian Patchworks has parameters that directly mirror feature importance. These parameters can be used to compare the importance of each feature for distinguishing individuals from different classes.

\begin{figure}[]
\centering
\includegraphics[scale=.35]{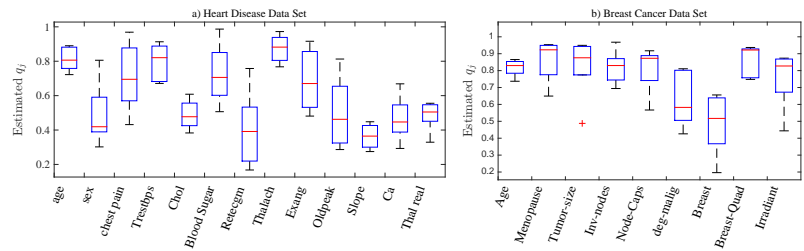}
 \caption{Posterior boxplots for $q_j$ associated with the features used in heart disease data set (left) and the features used in the breast cancer data set (right).}
\label{fig:posteriorbox1} 
\end{figure}

\begin{figure}[]
\centering
\includegraphics[scale=.5]{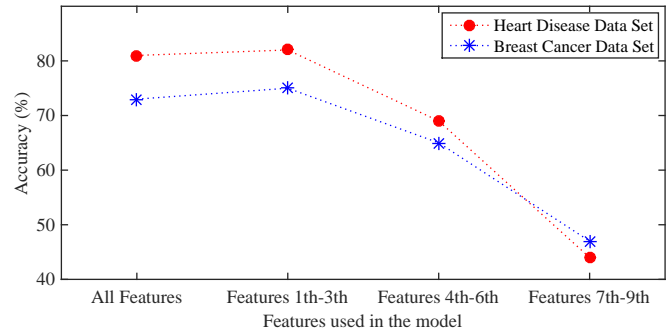}
 \caption{Accuracy evaluation based on ranked features for both data sets.}
\label{fig:posteriorbox2} 
\end{figure}

 In Fig. \ref{fig:posteriorbox1}, the posterior predictive distribution of $q_j$ for the features used in both data sets are shown for the unsupervised version of Model II. We can observe from this figure that features such as age and blood pressure seem to be more important than features like Serum Cholesterol or Resting Electrographic Results for finding the influential neighbors of each individual. We carried out the same experiments for the breast cancer dataset and plotted the posterior predictive distribution for $q_j$. 
Since we could not find any strong scientific resource to directly evaluate or approve these findings, we indirectly evaluate them, by comparing to another feature importance approach: 
first we used only the top 3 features to train the model; next, we used the next 3 features (feature ranked 4th, 5th, and 6th); finally the next 3 features (feature ranked 7th, 8th, and 9th) and evaluated the prediction power of the model and compared with the case of when all features are employed. Results shown in Fig. \ref{fig:posteriorbox2} indicate that for both datasets, the performance of the top three features alone is similar to when all features are used. When the less important features are employed, the prediction power of the model degrades. This result shows that the vector obtained from $q_j$ is actually indicative of feature importance. In Subsection \ref{subsec:knn}, we used estimated feature importance vectors as the feature weight vectors for KNN and saw slightly improved prediction performance. We repeated this using Model I and other extensions and the results were similar. 


\subsection{{
Sensitivity analysis for various hyperparameters}}\label{sec:sens}

Sensitivity analysis in a Bayesian setting can help find out how sensitive a model’s performance is to minor changes in its hyperparameters. We report below sensitivity analysis to hyperparameters $S$, $\alpha$, $\gamma$, $\sigma_1$, and $\sigma_2$.

\subsubsection{Sensitivity analysis for the number of parents ($S$)}
\begin{figure}[]
\centering
 \includegraphics[scale=.4,trim=1.5cm .0cm 1.5cm 0.cm]{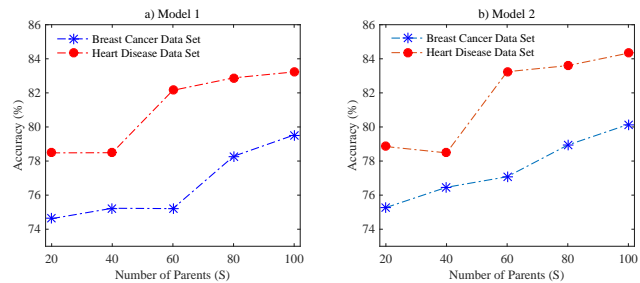}
 \caption{Prediction Accuracy of Model I and Model II under different values of S}\label{subsec:sensS}
\end{figure}

The total number of influential neighbors for each new case is a critical factor in the performance of the model. While increasing the number of parents $S$ will improve the coverage of parents among the distribution of patients, it can also make the computational time to train the model longer, and potentially lead to overfitting. 
In Figure \ref{subsec:sensS}, we show the accuracy of both Model I and Model II under different values of $S$. As can be seen for both data sets, the accuracy improves as we increase the number of cases in the parent set for our data. 

\subsubsection{Sensitivity analysis for $\alpha$}

\begin{figure}[]
\centering
 \includegraphics[scale=.4,trim=0.5cm .0cm 1cm 0.cm]{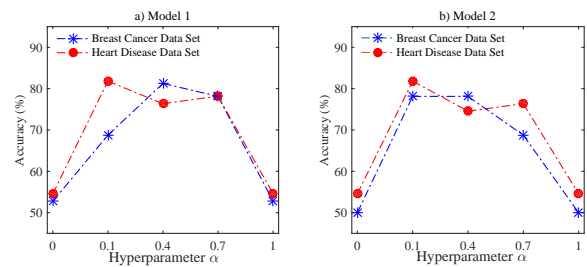}
 \caption{Prediction Accuracy of Model I and Model II under different values of $\alpha$}
 \label{fig:sensalpha}
\end{figure}

As explained earlier, parameter $\alpha$ controls how many parents are an influential neighbor of each individual. The average number of influential neighbors is $S \times \alpha$. It is expected that large $\alpha$ can improve the performance of the model, however choosing a very large $\alpha$ may result in selecting all parents as neighbors and possibly overfitting. Conversely, if $\alpha$ is chosen to be too small, then no parent is selected as the influential neighbor of each case and the model performs as random guessing. In Figure \ref{fig:sensalpha}, the accuracy of Model I and Model II under five different values of $\alpha={0, 0.1, 0.4, 0.7, 1}$ is given for the two data sets. 
Under the two extreme cases of $\alpha=0$ and $\alpha=1$, the model acts as a random guess with accuracy close to 50\%. The best choices for $\alpha$ are at non-extreme values; ideally $\alpha$ would be cross-validated if computation time permits.

\subsubsection{Sensitivity analysis for $(\gamma,\sigma_1)$}
\begin{figure}[]
\centering
 \includegraphics[scale=.4,trim=0.5cm .0cm 1cm 0.cm]{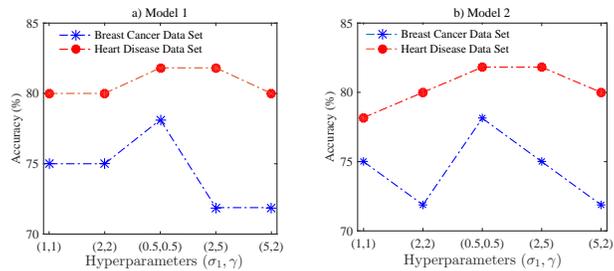}
 \caption{Prediction Accuracy of Model I and Model II under different values of $(\gamma,\sigma_1)$.
 \label{fig:sensitgamma}}
\end{figure}

As explained earlier, $q_j$ determines the importance of feature $j$ (ranging between 0-1) and follows a beta distribution with parameters $\gamma$ and $\sigma$. These hyperparameters will change the shape the distribution for $q_j$ and control the number of features and their importance. To explore different shape types of the beta distribution, we consider three symmetric cases: (i) a U-shaped distribution with $(\gamma,\sigma_1)$=(.5,.5), (ii) a uniform distribution with $(\gamma,\sigma_1)$=(1,1), (iii) a symmetric unimodal distribution with $(\gamma,\sigma_1)$=(2,2), and two unimodal cases (iv) $(\gamma,\sigma_1)$=(2,5) and (v) $(\gamma,\sigma_1)$=(5,2) with positive and negative skews, respectively. Figure \ref{fig:sensitgamma} shows the models' sensitivity to $(\gamma,\sigma_1)$. In both datasets, the symmetric case of $(\gamma,\sigma_1)=(0.5,0.5)$ (the Jeffreys prior for the Bernoulli and binomial distributions) performs best, but not by very much. 

We performed sensitivity analysis for other hyperparameters $\lambda_0$, $\lambda$, $\mu_0$, $\mu$, and $\sigma_2$ and found out that the model is not very sensitive within the range of reasonable choices.

\subsection{{
Comparison with Existing Models}}
\label{subsec:comp}



In this section, we first provide the results of a general comparison between our model and some existing models. We then more closely compare our model with the most similar models including BCM, KNN, and decision trees. Our purpose is not to show that our model outperforms these methods in terms of prediction; all methods perform comparably for our data. We aim instead to show that our model performs well enough, where interpretability serves as an excellent tie-breaker.


\subsubsection{Prediction Accuracy and Comparison with Benchmark Models}

The methods we compare with are logistic regression, SVM, BCM, AdaBoost, KNN, random forests, and decision trees. 
For each method and each dataset, we performed 5-fold cross-validation and reported the mean 
 over 5 folds for accuracy, sensitivity, specificity, precision, and recall. The results are summarized in Table \ref{tab:comparison1}, indicating that our methods (Model I and Model II) perform comparably with other machine learning algorithms.

\begin{table}[ht]
\centering
\caption{Results for comparison between our model (Model II) and selected existing methods on both data sets.}
\label{tab:comparison1}\resizebox{\textwidth}{!}{%
\begin{tabular}{@{}llllllllllllll@{}}
\toprule
\multicolumn{1}{c}{\textbf{Model}} & \multicolumn{6}{c}{\textbf{Heart Disease Dataset}} & & \multicolumn{6}{c}{\textbf{Breast Cancer Dataset}} \\ \midrule
\multicolumn{1}{c}{\textbf{Learning Type}} & \multicolumn{1}{c}{{Accuracy}} & \multicolumn{1}{c}{{Sensitivity}} & \multicolumn{1}{c}{Specificity} & \multicolumn{1}{c}{Precision} & \multicolumn{1}{c}{Recall} & \multicolumn{1}{c}{} & \multicolumn{1}{c}{} & \multicolumn{1}{c}{Accuracy} & \multicolumn{1}{c}{Sensitivity} & \multicolumn{1}{c}{Specificity} & \multicolumn{1}{c}{Precision} & \multicolumn{1}{c}{Recall} & \multicolumn{1}{c}{} \\ \midrule
Proposed Model I & 0.83 & 0.82 & 0.83 & 0.84 & 0.82 & & & 0.78 & 0.86 & 0.7 & 0.73 & 0.86 & \\
Proposed Model II & 0.84 & 0.84 & 0.84 & 0.84 & 0.84 & & & 0.79 & 0.87 & 0.71 & 0.75 & 0.87 & \\
BCM ($S= 4, \alpha=0.01$) & 0.80 &	0.86 & 0.74 &	0.78 &	0.86 & & & 0.76 & 0.93 & 0.60 & 0.69 & 0.93 & \\
 BCM ($S= 4, \alpha=1$) & 0.82 &	0.85 &	0.80 &	0.81 &	0.85 & & & 0.75 & 0.82 & 0.68 & 0.72 & 0.82 & \\
BCM ($S= 8, \alpha=0.01$) & 0.63 &	0.53 &	0.75 & 	0.64 &	0.53 & & & 0.68 & 0.97 & 0.40 & 0.62 & 0.97 & \\
BCM ($S= 8, \alpha=1$) & 0.78 &	0.80 &	0.77 &	0.78 &	0.80 & & & 0.77 & 0.84 & 0.71 & 0.75 & 0.84 & \\
logistic regression & 0.82 & 0.84 & 0.80 & 0.81 & 0.84 & & & 0.77 & 0.89 & 0.66 & 0.72 & 0.89 & \\
KNN (30)$^{*}$ & 0.83 & 0.85 & 0.82 & 0.83 & 0.85 & & & 0.76 & 0.99 & 0.54 & 0.68 & 0.99 & \\
KNN (20) & 0.82 & 0.86 & 0.79 & 0.80 & 0.86 & & & 0.76 & 0.96 & 0.56 & 0.69 & 0.96 & \\
KNN (10) & 0.81 & 0.82 & 0.80 & 0.81 & 0.82 & & & 0.75 & 0.99 & 0.53 & 0.68 & 0.99 & \\
KNN (5) & 0.81 & 0.82 & 0.81 & 0.81 & 0.82 & & & 0.73 & 0.90 & 0.57 & 0.67 & 0.90 & \\
KNN (1) & 0.76 & 0.74 & 0.78 & 0.77 & 0.74 & & & 0.68 & 0.70 & 0.65 & 0.66 & 0.70 & \\
Random Forest (5)$^{**}$ & 0.78 & 0.79 & 0.77 & 0.78 & 0.79 & & & 0.78 & 0.85 & 0.72 & 0.75 & 0.85 & \\
Random Forest (10) & 0.80 & 0.81 & 0.79 & 0.80 & 0.81 & & & 0.74 & 0.80 & 0.68 & 0.71 & 0.80 & \\
Random Forest (100) & 0.83 & 0.84 & 0.82 & 0.83 & 0.84 & & & 0.78 & 0.87 & 0.70 & 0.74 & 0.87 & \\
Decision Tree & 0.75 & 0.77 & 0.72 & 0.76 & 0.77 & & & 0.73 & 0.70 & 0.75 & 0.73 & 0.70 & \\
SVM & 0.81 & 0.85 & 0.76 & 0.79 & 0.85 & & & 0.79 & 1.00 & 0.59 & 0.71 & 1.00 & \\
boosted DT (100, I)$^{***}$ & 0.82 & 0.80 & 0.84 & 0.83 & 0.80 & & & 0.77 & 0.90 & 0.65 & 0.72 & 0.90 & \\
boosted DT (50, I) & 0.82 & 0.81 & 0.84 & 0.84 & 0.81 & & & 0.79 & 0.93 & 0.65 & 0.73 & 0.93 & \\
boosted DT (10, I) & 0.83 & 0.81 & 0.84 & 0.84 & 0.81 & & & 0.79 & 0.99 & 0.60 & 0.71 & 0.99 & \\
boosted DT (5, I) & 0.80 & 0.81 & 0.78 & 0.80 & 0.81 & & & 0.76 & 0.91 & 0.62 & 0.70 & 0.91 & \\
boosted DT (100, II) & 0.81 & 0.85 & 0.77 & 0.79 & 0.85 & & & 0.77 & 0.90 & 0.65 & 0.72 & 0.90 & \\
boosted DT (50,II) & 0.81 & 0.85 & 0.77 & 0.79 & 0.85 & & & 0.77 & 0.90 & 0.65 & 0.72 & 0.90 & \\
boosted DT (10, II) & 0.81 & 0.85 & 0.77 & 0.79 & 0.85 & & & 0.78 & 0.91 & 0.66 & 0.73 & 0.91 & \\
boosted DT (5, II) & 0.81 & 0.85 & 0.77 & 0.79 & 0.85 & & & 0.78 & 0.91 & 0.65 & 0.72 & 0.91 & \\ \bottomrule
\end{tabular}}
\small
\sloppy
\begin{flushleft}
\small
$^{**}$The numbers in the parentheses are the number of parents. The distance measure for KNN is Hamming. $^{***}$The number in the parentheses for random forests is the number of trees.
$^*$The numbers in the parentheses are the number of the weak classifiers and the weak learning algorithm type for AdaBoost. The learning for Type 1 is Tree and for Type II is Discriminant. We used the fitensemble function in Matlab with AdaBoostM1 as the method. 
\end{flushleft}
\normalsize
\end{table}

\subsubsection{Comparison with KNN and its extensions}
\label{subsec:knn}

\begin{figure}[ht]
 \includegraphics[scale=.4,trim=2cm .0cm 1cm 0.cm]{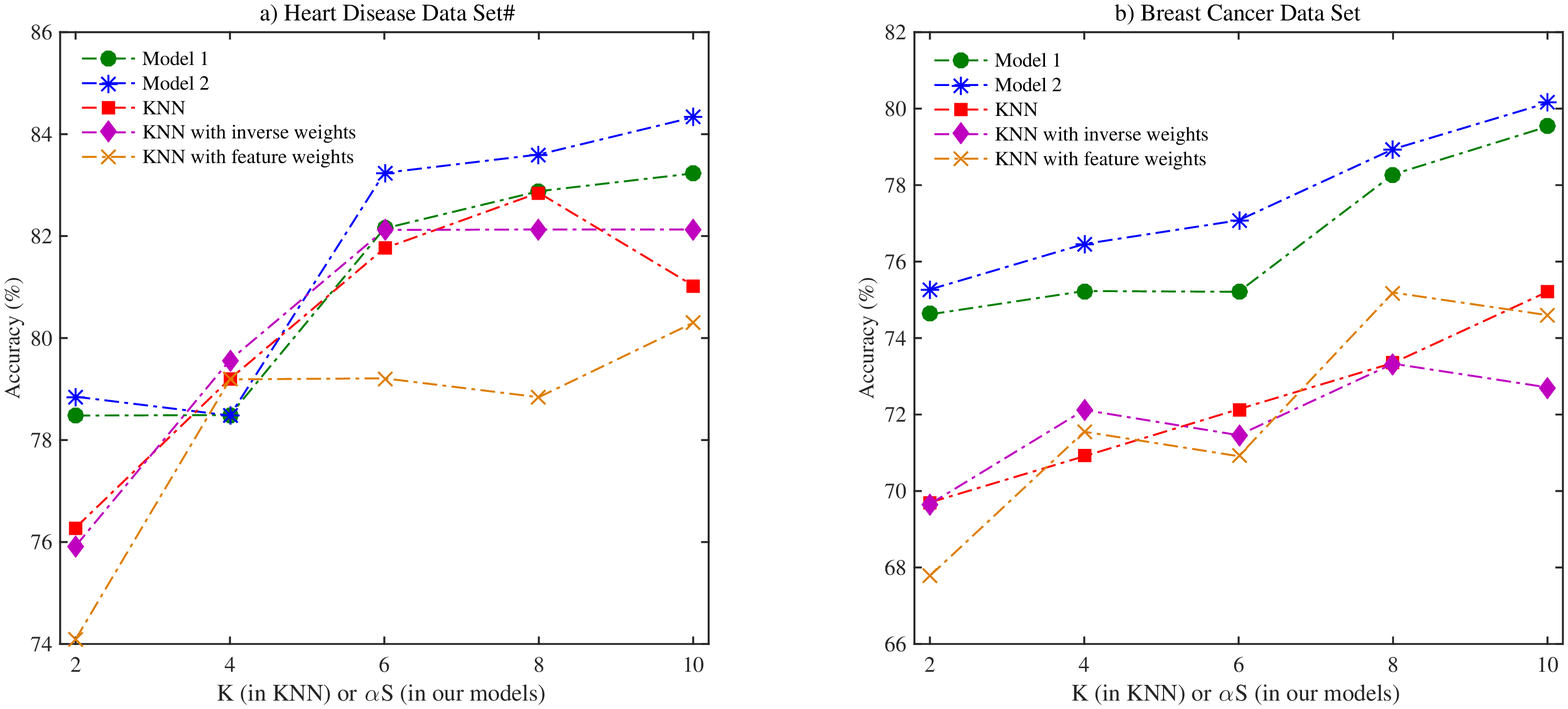}
 \caption{Comparison between our model, KNN, and its two extensions.}
 \label{fig:comp3}
\end{figure}

One of the main differences between Bayesian Patchworks and the regular KNN is that unlike KNN that requires a distance measure that is the same for all instances of data, our model uses an adaptive metric, where it finds neighbors from parents based on their important features. 
Also, unlike KNN, BPatch does not force a fix number of neighbors/parents for each case. 
In our experiments, the neighbors chosen by KNN were not always similar to those chosen by BPatch.
There are some extensions of KNN that have addressed some of the limitations of KNN (see the survey for such techniques \citep{Bhatia2010}). As explained in \citep{Bhatia2010}, most of these extensions are improvements over basic kNN to gain speed efficiency as well as space efficiency and may hold good in particular field under particular circumstances. 
Since we cannot compare our model with all of the extensions of KNN, we chose two of the most relevant extensions of KNN and compare our results with them. The first extension is the KNN with distance weighted votes, where the training points are assigned weights according to their distances from sample data point (we used the inverse of the distance). The second extension is KNN with feature weights. We chose $K$ in KNN and $S\alpha$ in our models to be the same, so that, on average, the same number of features are selected for each neighbor. The results of this comparison are shown in Fig. \ref{fig:comp3}. The performance of our model was still better than KNN and its two extensions. 
In KNN and its extensions, whether or not distance weights, feature weights, or bandwidths are considered, the comparison between a new case and a case in the training set is always based on a fixed formula with a fixed number of neighbors. Our distance metric is more flexible.

\subsubsection{Comparison with BCM}

\begin{figure}[]
\centering
 \includegraphics[scale=.5,trim=2cm .0cm .cm 0.5cm]{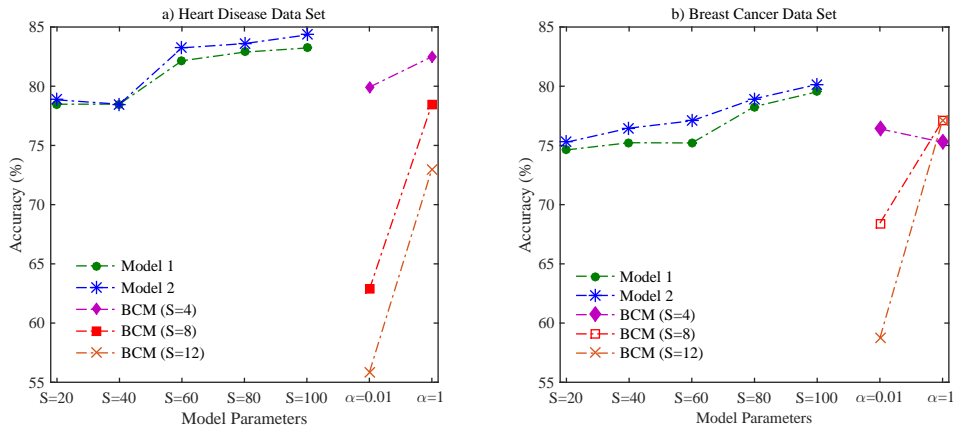}
 \caption{Comparison between the proposed model and BCM.}
 \label{fig:bcmcompare}
\end{figure}

To adapt the Bayesian Case Model \cite{KimRuSh14} for supervised classification (and health outcome prediction), the mixture vector $\bm{\pi}$ from BCM can be used as a feature of length $S$ for input to other classification methods, such as SVM and logistic regression. 
 In Fig. \ref{fig:bcmcompare}, we compared our models (Model I and Model II) with four possible values for $S\in\{20,40,60,80,100\}$ and compared it with 6 cases of BCM (combinations of $S=4, S=8, S=12$, and $\alpha=0.01$ and $\alpha=.1$). Note that no larger $S$ was considered for BCM simply due to the low performance of the trained models. The performance of BPatch was generally better.

Interpreting BPatch's results may also be easier than interpreting those of BCM.
The posterior estimate of the mixture vector $\bm{\pi}$ over the prototypes is not directly interpretable since the prototypes are not fixed, but we can make choices that allow for sparser or larger mixtures. Let us consider BCM under two conditions for the hyperparameter $\alpha$ used in \cite{KimRuSh14}, which controls the sparsity of the prototypes chosen for a new case. 

\noindent \textbf{{Sparse BCM (e.g., $\alpha=0.01$):}} Only one prototype generates each observation. Table \ref{fig:sample1} shows this prototype, which shares some of the feature values with the new observation.


\noindent
\textbf{Non-sparse BCM (e.g., $\alpha=1$)}: BCM chooses multiple prototypes, however, \textit{only one} prototype generates each feature of the new patient. In contrast, BPatch's features are generated from multiple parents. Table \ref{fig:sample1} illustrates this.


\begin{table}[ht]
\centering
\caption{A sample patient, predicted prototypes in BCM (sparse \& non-sparse) vs. parents in our model (Model II)}
\label{fig:sample1}\resizebox{\textwidth}{!}{
\begin{tabular}{@{}clcccllllllllllcc@{}}
\toprule
\multirow{2}{*}{Model} & Feature & F1 & F2 & F3 & F4 & F5 & F6 & F7 & F8 & F9 & F10 & F11 & F12 & F13 & \begin{tabular}[c]{@{}c@{}}Similarity \\ Index\end{tabular} & \begin{tabular}[c]{@{}c@{}}True \\ Health\end{tabular} \\ \cmidrule(l){2-17} 
 & Sample Patient & {{2}} & 1 & 3 & 2 & 3 & 1 & 1 & 3 & 2 & 2 & 2 & 1 & 2 & & 2 \\ \midrule
\begin{tabular}[c]{@{}c@{}}BCM\\ $\alpha$=.01\end{tabular} & 1st Prototype & \underline{\textbf{3}} & \underline{\textbf{2}} & \underline{\textbf{3}} & \underline{\textbf{3}} & \underline{\textbf{1}} & \underline{\textbf{2}} & \underline{\textbf{2}} & \underline{\textbf{3}} & \underline{\textbf{1}} & \underline{\textbf{2}} & \underline{\textbf{2}} & \underline{\textbf{2}} & \underline{\textbf{2}} & \underline{\textbf{1}} &
	\underline{\textbf{2}} \\ \midrule
\multirow{4}{*}{\begin{tabular}[c]{@{}c@{}}BCM\\ $\alpha$=1\end{tabular}} & 1st Prototype & \underline{\textbf{2}} & \underline{\textbf{1}} & 2 & 2 & \underline{\textbf{3}} & 2 & 1 & {{3}} & 1 & 1 & 1 & \underline{\textbf{1}} & 1 & 0.15 & 1 \\
 & 2nd Prototype & 3 & 2 & {{3}} & 3 & 3 & \underline{\textbf{1}} & \underline{\textbf{1}} & 3 & \underline{\textbf{2}} & {{2}} & \underline{\textbf{2}} & 2 & {{2}} & 0.54 & 2 \\
 & 3rd Prototype & 1 & 2 & 3 & \underline{\textbf{2}} & 1 & 1 & 1 & 3 & 1 & 1 & 1 & 1 & 1 & 0.00 & 1 \\
 & 4th Prototype & 1 & 2 & \underline{\textbf{3}} & 1 & 1 & 1 & 1 & \underline{\textbf{3}} & 2 & 2 & \underline{\textbf{2}} & 1 & \underline{\textbf{2}} & 0.31 & {{2}} \\ \midrule
\multirow{4}{*}{Model II} & 1st Parent & \underline{\textbf{2}} & \underline{\textbf{1}} & \underline{\textbf{3}} & \underline{\textbf{2}} & \underline{\textbf{3}} & \underline{\textbf{1}} & \underline{\textbf{1}} & \underline{\textbf{3}} & \underline{\textbf{2}} & \underline{\textbf{2}} & \underline{\textbf{2}} & \underline{\textbf{1}} & \underline{\textbf{2}} & 1 & 2 \\
 & 2nd Parent & 3 & \underline{\textbf{1}} & \underline{\textbf{3}} & 3 & \underline{\textbf{3}} & \underline{\textbf{1}} & 2 & 3 & 2 & 2 & 2 & \underline{\textbf{1}} & \underline{\textbf{2}} & 0.11 & 2 \\
 & 3rd Parent & 3 & \underline{\textbf{1}} & \underline{\textbf{3}} & 3 & \underline{\textbf{3}} & \underline{\textbf{1}} & 2 & 3 & 2 & 2 & 2 & \underline{\textbf{1}} & 1 & 0.08 & 2 \\
 & 4th Parent & 2 & \underline{\textbf{1}} & 2 & 3 & 1 & \underline{\textbf{1}} & \underline{\textbf{2}} & 3 & 2 & 2 & 2 & \underline{\textbf{1}} & 1 & 0.05 & 1 \\ \bottomrule
\end{tabular}}
\end{table}


\subsubsection{Comparison with Decision Trees}

Decision trees are a completely different form of interpretable model that is not case-based. In the datasets we are considering, there are many good predictive models (a large ``Rashomon Set") meaning that many models that are different from each other that all perform well. We used CART, 
whose performance (shown in Table \ref{tab:comparison1}) was not as strong as some of the other methods. Interestingly, the features that CART identifies were not the same as those identified by BPatch. The feature ranking from both methods is shown in Fig. \ref{fig:comd31}.
More result is given in Appendix \ref{sec:appD}.

\begin{figure}[ht]
\centering
 \includegraphics[scale=.35,trim=3cm .0cm 1cm 0.cm]{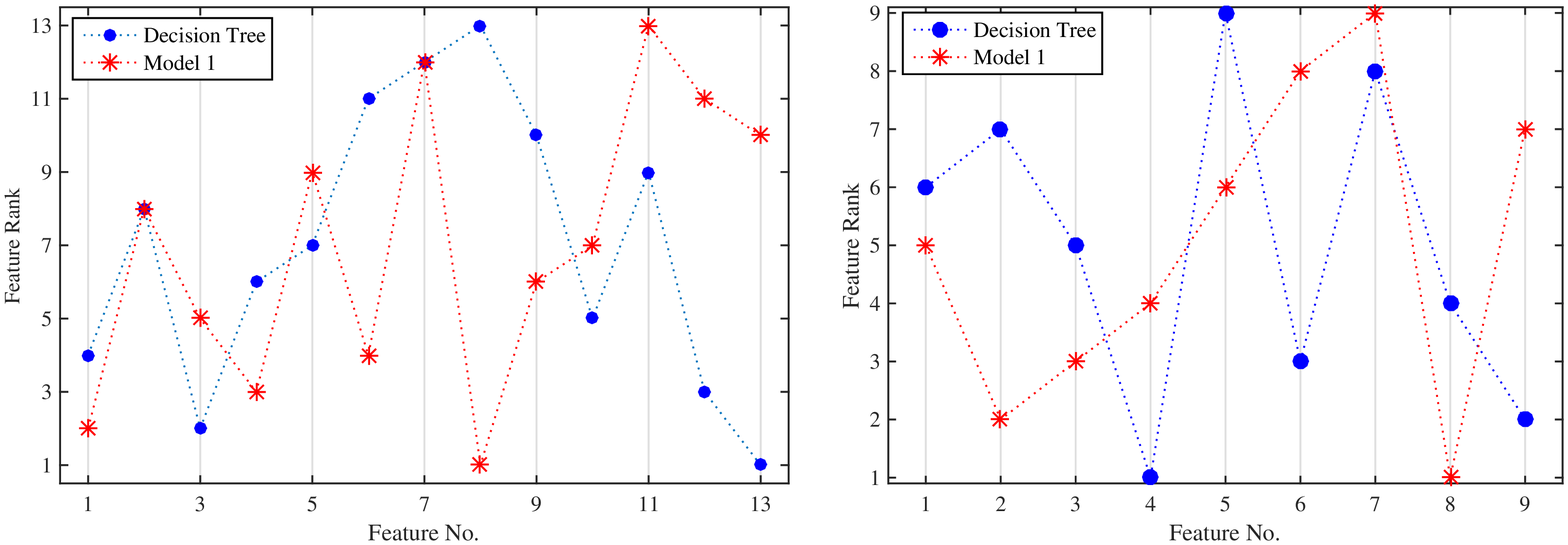}
 \caption{Feature ranking comparison between BPatch and decision trees (left: heart disease data set, right: breast cancer data set.}
 \label{fig:comd31}
\end{figure}

\subsection{Application to Larger Dataset and its Computational Challenges}\label{sec:large}

Computation is the major challenge for BPatch. Here, we show that BPatch's performance does not deteriorate on a slightly larger dataset with more samples and more features, despite more challenging computation.  

This dataset is provided by the Center for Clinical and Translational Research, Virginia Commonwealth University and is also available at the UCI website \citep{Strack2014}. This data has been prepared to analyze factors related to readmission as well as other outcomes pertaining to patients with diabetes. The dataset represents 10 years (1999-2008) of clinical care at 130 US hospitals and integrated delivery networks. 
 Similar to \cite{Strack2014}, we focused on early readmission, and defined the readmission label (i.e., health outcome) as having two values: ``readmitted,” if the patient was readmitted within 30 days of discharge or ``otherwise,” which covers both readmission after 30 days and no readmission at all. 
 The list of the features and their possible outcomes are given in Table \ref{tab:read}, where features age and time-in hospital were discretized. We used data from two sets of 500 patients each, performed 5-fold cross-validation, and reported the average accuracy for BPatch and similar methods in Table \ref{tab:readres}. The hyperparameters are the same as previous experiments and the number of parents was set to 60. 
 Again, most methods give comparable results, with BPatch's accuracy somewhere in the middle. For interpretability, we also checked (not shown here) that selected neighbors were reasonably close to their respective patients.

\begin{table}[ht]
\centering
\caption{The attributes used in this paper for the readmission dataset}
\label{tab:read}
\resizebox{\textwidth}{!}{%
\begin{tabular}{@{}lllll@{}}
\toprule
\multicolumn{1}{c}{feature Name} & \multicolumn{1}{c}{Description} & & \multicolumn{1}{c}{feature Name} & \multicolumn{1}{c}{Description} \\ \midrule
Race & \begin{tabular}[c]{@{}l@{}}Caucasian, AfricanAmerican, \\ others\end{tabular} & & Final diagnosis & \begin{tabular}[c]{@{}l@{}}Diabetes, circulatory system, \\ respiratory system, digestive system, \\ Injury and poisoning, Other\end{tabular} \\
Gender & male, female & & Metformin & Up, Down \\
Age & {[}10(i-1)--10i{]}, $i \in$ \{1..9\} & & Glimepiride & Up, Down \\
discharge\_disposition\_id & Discharged to home, otherwise & & Glipizide & Up, Down \\
Admission\_source & \begin{tabular}[c]{@{}l@{}}emergency room, physician/clinic \\ referral, otherwise\end{tabular} & & Glyburide & Up, Down \\
Time\_in\_hospital & \textless2,2-7, \textgreater7 & & Pioglitazone & Up, Down \\
Medical\_specialty & \begin{tabular}[c]{@{}l@{}}InternalMedicine, Family/General \\ Practice, Cardiology, others\end{tabular} & & Rosiglitazone & Up, Down \\
A1Cresult & \textgreater7, \textgreater 8, none, other & & Insulin & Up, Down \\
DiabetesMed & Yes, No & & Admission\_type & Emergency, Other \\
Changes in diabetes & Yes, no & & & \\ \bottomrule
\end{tabular}}
\end{table}

\begin{table}[ht]
\centering
\caption{Results for comparison between our model and benchmark models in the readmission data sets}
\label{tab:readres}
\resizebox{\textwidth}{!}{
\begin{tabular}{@{}lccccccc@{}}
\toprule
Learning Type & Accuracy & Sensitivity & Specificity & Precision & Recall & F-measure \\ \midrule
Model (I) - ($S$=60) & 74.50 &	84.97 &	64.05 &	70.58 &	84.97 &	79.26 \\
Our Model (II)- ($S$=60) & 75.4 & 82.7 & 68.0 & 72.6 & 82.7 & 79.1 \\
BCM ($S=4, \alpha=1$) & 76.80 &	85.99 &	67.73 &	72.73 &	85.99 &	78.74 \\
BCM ($S=8, \alpha=1$) & 77.30 &	87.40 &	67.30 &	72.79 &	87.40 &	79.36 \\
Logistic Regression & 75.9 & 78.5 & 73.2 & 74.5 & 78.5 & 76.4 \\
KNN (30) & 62.3 & 96.1 & 28.7 & 57.5 & 96.1 & 71.8 \\
KNN (20) & 63.1 & 96.5 & 29.8 & 58.0 & 96.5 & 72.4 \\
KNN (10) & 64.1 & 95.4 & 32.9 & 59.0 & 95.4 & 72.7 \\
KNN (5) & 67.3 & 92.7 & 42.3 & 61.8 & 92.7 & 73.9 \\
KNN (1) & 66.9 & 88.0 & 45.9 & 62.0 & 88.0 & 72.6 \\
Random Forest (5)& 78.8 & 83.2 & 74.4 & 76.7 & 83.2 & 79.6 \\
Random Forest (10) & 80.6 & 84.6 & 76.5 & 78.5 & 84.6 & 81.4 \\
Random Forest (100) & 83.1 & 85.7 & 80.5 & 81.7 & 85.7 & 83.6 \\
Decision Tree & 77.9 & 78.9 & 76.9 & 77.5 & 78.9 & 78.1 \\
SVM & 67.0 & 88.2 & 45.6 & 65.6 & 88.2 & 73.6 \\
boosted DT (100, I) & 79.3 & 82.4 & 75.9 & 77.9 & 82.4 & 80.0 \\
boosted DT (50, I) & 78.7 & 83.0 & 74.2 & 76.7 & 83.0 & 79.6 \\
boosted DT (10, I) & 75.0 & 82.6 & 67.5 & 72.3 & 82.6 & 76.7 \\
boosted DT (5, I) & 72.4 & 84.7 & 60.0 & 68.5 & 84.7 & 75.3 \\
boosted DT ( 100, II) & 74.3 & 79.7 & 68.7 & 72.1 & 79.7 & 75.5 \\
boosted DT (50, II) & 74.3 & 79.7 & 68.7 & 72.1 & 79.7 & 75.5 \\
boosted DT (10, II) & 74.3 & 79.7 & 68.7 & 72.1 & 79.7 & 75.5 \\
boosted DT (5, II) & 74.3 & 79.7 & 68.7 & 72.1 & 79.7 & 75.5 \\ \bottomrule
\end{tabular}}
\end{table}

\begin{figure}[]
\centering
 \includegraphics[scale=.35,trim=0.0cm .0cm 1cm 0.cm]{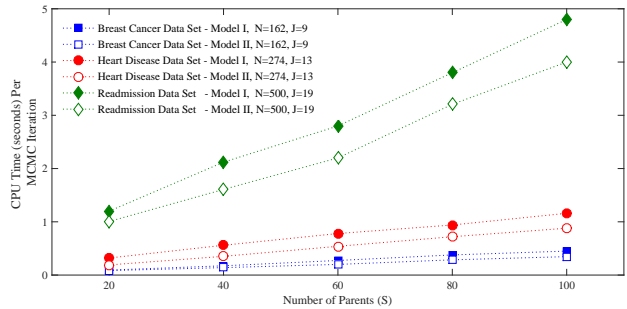}
 \caption{CPU comparison between Model I and Model $2$ and their sensitivity to $S$ and $J$}
 \label{fig:cpu1}
\end{figure}

As shown in Fig. \ref{fig:cpu1} for all three data sets, as the number of parents increases, it takes more time to learn the structure of the model. 
It is also shown that training Model I is faster than Model II. 
Also, given the same number of parents and prototypes, each step of BPatch and BCM takes around the same amount of time, but BPatch uses more memory, mainly due to the existence of 3-dimensional variable $\bm{w}$. 
Thus, we can conclude that without sacrificing accuracy, our model is slightly slower than BCM and much slower than the rest of the models presented in Table \ref{tab:comparison1}. On the other hand, there is something to be gained from that extra computation, and the tradeoff between computation and the need for interpretability should depend on the application. BPatch is currently suited for smaller datasets (e.g., clinical trials, data from rare diseases, preliminary studies with high-quality hand-curated samples, data from one physician's case files, or data from one department in a hospital) where one requires a high level of interpretability in order to reason about cases.

\section{Conclusion and Future Work}
\label{SecConc}
Doctors reason through cases already. In a sense, we are proposing a data-driven formalization of techniques that are used in an ad hoc way throughout much of current practice.
In some sense, one could view Bayesian Patchworks as a form of adaptive $k$ nearest neighbors. Each case is generated by several neighbors, each using a different distance measure to the present case. In the generative sense, each patient is built the way a patchwork quilt is built; the larger design is built from pieces gathered from other sources. It is easy to argue that considering neighbors through different lenses (as Bayesian Patchworks does) makes more sense than fixing one distance measure, and having only one way to search for nearest neighbors. 

Bayesian Patchworks operates similarly to real estate ``comps," which are similar properties previously sold that one might use to price a current house on the market. There are many possible future extensions of Bayesian Patchworks, for instance, one could modify the prior to favor using fewer total parents. Or, one could generalize the method to handle problems other than classification. Regression would be a trivial extension, but more complex outputs could be more interesting.



%
%
%
%

\ifCLASSOPTIONcompsoc
 \section*{Acknowledgments}
\else
 \section*{Acknowledgment}
\fi
The authors would like to thank Siemens and Wistron for supporting this work.
\ifCLASSOPTIONcaptionsoff
 
\fi

\appendices 

\vspace{-.1in}
\section{Appendix: Inference}\label{SecInference}
 The model hyperparameters are denoted as $\bm{\vartheta}=\{\alpha, \gamma, \sigma_1, \sigma_2, \lambda_0, \lambda, \mu_0, \mu\}$. Given a set of known $(\mathbf{X},\mathbf{Y})$ and hyperparameters $\bm{\vartheta}$, the full posterior, which is the product of the likelihood and priors, can be computed as
\begin{flalign} 
	& \Pr(\bm{q},\bm{\tilde{q}},\bm{z},\bm{w},\bm{\phi},\bm{\theta}|\mathbf{X},\mathbf{Y},\bm{\vartheta}) \propto 
\Pr(\mathbf{X},\mathbf{Y}|\bm{q},\bm{\tilde{q}},\bm{z},\bm{w},\bm{\phi},\bm{\theta},\bm{\vartheta}) \times \Pr(\bm{q},\bm{\tilde{q}},\bm{z},\bm{w},\bm{\phi},\bm{\theta}|\bm{\vartheta}) \\ & =\nonumber
\Pr(\mathbf{Y}|\bm{\theta},\bm{\vartheta}) \times \Pr(\mathbf{X}|\bm{\phi},\bm{\vartheta}) \\ \nonumber & \times 
\Pr(\bm{\phi}|\bm{w},\bm{z},\bm{\vartheta}) \times
\Pr(\bm{\theta}|\bm{z},\bm{\vartheta}) \times \Pr(\bm{z}|\bm{\vartheta}) \times \Pr(\bm{w}|\bm{\tilde{q}},\bm{\vartheta}) \times \Pr(\bm{\tilde{q}}|\bm{q},\bm{\vartheta}) \times \Pr(\bm{q},\bm{\vartheta}).
\end{flalign}
We can use conjugate priors to simplify the above expression and also integrate out the latent factors $\bm{\phi},\bm{\theta}, \text{and}, \bm{\tilde{q}}$. This simplifies to: 
\begin{flalign}	\label{eq:postint2}
&\int \int \int \Pr(\bm{q},\bm{\tilde{q}},\bm{z},\bm{w},\bm{\phi},\bm{\theta}|\mathbf{X},\mathbf{Y},\bm{\vartheta}) \, d\bm{\tilde{q}}\, d\bm{\theta} \, d\bm{\phi} & 
\\ \nonumber 
&\propto \int \Pr(\bm{w}|\bm{\tilde{q}},\bm{\vartheta}) \Pr(\bm{\tilde{q}}|\bm{q},\bm{\vartheta}) \, d\bm{\tilde{q}} & &  (=\Pr(\bm{w}|\bm{q},\bm{\vartheta}))
\\ \nonumber & \times \int \Pr(\mathbf{X}|\bm{\phi},\bm{\vartheta}) \Pr(\bm{\phi}|\bm{w},\bm{z},\bm{\vartheta}) d\bm{\phi} & &  (=\Pr(\mathbf{X}|\bm{z},\bm{w},\bm{\vartheta}))
\\ \nonumber & \times \int \Pr(\mathbf{Y}|\bm{\theta},\bm{\vartheta}) \Pr(\bm{\theta}|\bm{z},\bm{\vartheta}) d\bm{\theta} & & (=\Pr(\mathbf{Y}|\bm{z},\bm{\vartheta}))
\\ \nonumber & \times \Pr(\bm{q}|\bm{\vartheta}) \times \Pr(\bm{z}|\bm{\vartheta}) \\ \nonumber & =
\Pr(\bm{w}|\bm{q},\bm{\vartheta}) \times \Pr(\mathbf{X}|\bm{z},\bm{w},\bm{\vartheta}) \times \Pr(\mathbf{Y}|\bm{z},\vartheta) \times \Pr(\bm{q}|\bm{\vartheta}) \times \Pr(\bm{z}|\bm{\vartheta}). 
\end{flalign}
The rest of the calculations will be for the three integrations inside Eq. \eqref{eq:postint2} (lines 2-4). After performing these integrations, we will be left only with $\bm{q}$, $\bm{z}$, and $\bm{w}$. The details of those calculations are given in Appendix \ref{app:details}. From the results given in Appendix \ref{app:details}, we have the following three important relationships, which can be used to compute 
$\Pr(\bm{w}|\bm{q},\bm{\vartheta})$, $ \Pr(\mathbf{X}|\bm{z},\bm{w},\bm{\vartheta})$, and $\Pr(\mathbf{Y}|\bm{z},\vartheta)$ in Eq. \eqref{eq:postint2}, respectively: 
\begin{flalign}\label{eq:simpw}
	 \Pr(w_{:bj}|q_{j}) & = \int_0^1 \left( \Pr(w_{:bj}|\tilde{q}_{bj})\right) \Pr(\tilde{q}_{bj}|q_j) \, d{\tilde{q}_{bj}}
\\ \nonumber &=
 \frac {\mathbf{B}(q'_j+\sum \limits_i \mathbf{1}\{w_{ibj}=1\},\sigma_2+N - \sum \limits_i \mathbf{1}\{w_{ibj}=1\})} 
{\mathbf{B}(q'_j,\sigma_2)},
\end{flalign}

\begin{flalign}\label{eq:xphi}
 \Pr(x_{ij}|w_{i:j},z_{i:}) = \int \Pr(x_{ij}|\bm{\phi}_{ij}) \Pr(\bm{\phi}_{ij}|w_{i:j},z_{i:}) d\bm{\phi}_{ij} = \frac{g_{ij(x_{ij})}+1}{\sum \limits_v g_{ij}(v)+1}.
\end{flalign}

\begin{flalign}\label{eq:simyphi}
\Pr({y}_i|z_{i:}) = \int \Pr({y}_i|\bm{\theta}_i) \Pr(\bm{\theta}_i|z_{i:}) d\bm{\theta}_i = \frac{h_{i}(y_{i})+1}{\sum \limits_m h_{i}(m)+1}.
\end{flalign}
where $q'_j=\sigma_2 q_j/(1-q_j)$ and $\mathbf{B}(a,b)$ is the Beta function with parameters a and $b$.

\subsection{Metropolis-within-Gibbs (MWG)}\label{sec:MWG}

An iteration of MCMC cycles over the components of the parameter vectors $\bm{q}$, $\bm{z}$, and $\bm{w}$, keeping everything else fixed. 
 We first show that the conditional posterior distributions of $\bm{w}$, $\bm{z}$, and $\bm{q}$ admit closed forms. Denoting $q_j^-$ as the set of all parameters in the model except for $q_j$ (including the $w$'s and $z$'s, and implicitly including the variables that have been integrated out already), we have 
\begin{flalign}\label{eq:qupdate}
\Pr(q_j|q_j^-) &\propto	\Pr(q_j) \times \prod_b \Pr(w_{:bj}|q_j)\\ &=\nonumber \Pr(q_j) \times \prod_b \frac{\textbf{B}(q'_j+\sum \limits_i \mathbf{1}\{w_{ibj}=1\},\sigma_2+N - \sum \limits_i \mathbf{1}\{w_{ibj}=1\})} {\textbf{B}(q'_j,\sigma_2)}, 
\end{flalign}
where $\prod \limits_i \Pr(w_{ibj}|q_j)$ is obtained from Eq. \eqref{eq:simpw}. The step above uses conditional independence with respect to the $q_j$'s, and other conditional independences specified in the model. It should be noted that given a node's parents in a directed acyclic graph, that node is conditionally independent of its grandparents and any other ancestors (see Proposition 5.2 in \cite{jackman2009bayesian} on how to find conditional distribution using directed acyclic graph).
Now we have a simple form for the conditionals that drive the MCMC sampler. We use a Metropolitan Hastings algorithm to sample from the above conditional distribution. 

The rest of the parameters ($\bm{z}$ and $\bm{w}$) in the model can be sampled directly from the conditional distributions using Gibbs sampling. In particular, using $\prod \limits_i \Pr(w_{ibj}|q_j)$ from Eq. \eqref{eq:simpw} and $\Pr(x_{ij}|w_{i:j},z_{i:})$ from Eq. \eqref{eq:xphi}, we get the following: 
\begin{flalign}\label{eq:wupdate}
&\Pr(w_{ibj}=k|w_{ibj}^-) 
 \propto \Pr(w_{ibj}=k|q_j) \times \Pr(x_{ij}|w_{i:j},z_{i:},w_{ibj}=k) \\ \nonumber
	& \propto \frac{\textbf{B}(q'_j+\sum \limits_{i'} \mathbf{1}\{w_{{i'}bj}=1\},\sigma_2+N - \sum \limits_{i'} \mathbf{1}\{w_{{i'}bj}=1\})} {\textbf{B}(q'_j,\sigma_2)} \times \frac{g_{ij(x_{ij})}+1}{\sum \limits_v g_{ij}(v)+1}, (\text{given}\,\, w_{ibj}=k),
\end{flalign}
where $w_{ibj}^-$ includes all variables except $w_{ibj}$. This means that we can sample directly from the above conditional distribution as $k$ takes only two possible values of 0 and 1. 

Finally, using that the $x_{ij}$'s and $y_i$'s are the children of $z_{ib}$, we get
\begin{flalign}
\Pr(z_{ib}=k|z_{ib}^-) \propto \Pr(z_{ib}=k) \times \prod_j \Pr(x_{ij}|w_{i:j},z_{i:}) \times \Pr({y}_i|z_{i:}), \end{flalign}
which can be simplified using Eqs. \eqref{eq:xphi} and \eqref{eq:simyphi} to
\begin{equation}\label{eq:zupdate}
\Pr(z_{ib}=k|z_{ib}^-) \propto \left\{ \begin{array}{ll}
 (1-\alpha) \prod \limits_j \frac{g_{ij}(x_{ij})+1}{\sum \limits_v g_{ij}(v)+1}
 \frac{h_{i}(y_{i})+1}{\sum \limits_m h_{i}(m)+1}, (\text{given}\,\, z_{ib}=0)
 & \mbox{if $k=0$};\\
 \alpha \prod \limits_j \frac{g_{ij}(x_{ij})+1 }{\sum \limits_v g_{ij}(v)+1}
 \frac{h_{i}(y_{i})+1}{\sum \limits_m h_{i}(m)+1}, (\text{given}\,\, z_{ib}=1) & \mbox{if $k=1$}.\end{array} \right.
\end{equation}
Now, we can implement a Metropolis within-Gibbs sampling method to generate samples of the posterior distribution. After we draw samples for $
\bm{z}=\{z_{ib}\}$ (Eq. \ref{eq:zupdate}), $\bm{q}=\{q_{j}\}$ (Eq. \ref{eq:qupdate}), and $\bm{w}=\{w_{ibj}\}$ (Eq. \ref{eq:wupdate}), we then calculate the posterior predictive distribution for all measures of interest. At iteration $t$, we sample

1) $\bm{z}^{(t)}$ from $\Pr(z_{ib}|z_{ib}^{-} \,^{(t)})$ using direct sampling, 

2) $\bm{q}^{(t)}$ from $\Pr(q_j|q_j^{-}\, ^{(t)})$ using random-walk Metropolis-Hastings, and

3) $\bm{w}^{(t)}$ from $\Pr(w_{ibj}|w_{ibj}^{-}\, ^{(t)})$ using direct sampling.

\noindent For better identifiably, we can add a constraint for $\bm{w}$ so that if $z_{ib}=0$, then all $w_{ibj}$ become zero for all $j$. However, since this changes the structure of the directed acyclic graph and adds more steps to calculate the conditional distributions for $\bm{z}$ and $\bm{w}$, we chose not to do that in this paper. 

\vspace{-.1in}
\subsection{Inference for outcomes}

Let us assume that $\bm{\Psi}=\{\bm{z},\bm{w},\bm{q}\}$ is the set of parameters that can be learned from dataset ($ \mathbf{X},\mathbf{Y}$), which includes $N+S$ past cases which do not include patient $r$. Similarly, let $\bm{\Psi}_r$ denote the parameters associated only with the $r$th individual (i.e., $\bm{\Psi}_r=\{z_{r:},w_{r::}\}$). We can write the posterior predictive distribution as:
\begin{flalign}
	\Pr(y_r=m|{\bm{x}_r},\mathbf{X},\mathbf{Y})= \int_{(\bm{\Psi}, \bm{\Psi}_r)} \Pr (y_r=m|\bm{\Psi},\bm{\Psi}_r) \Pr(\bm{\Psi}_r,\bm{\Psi}|\mathbf{X},\mathbf{Y},{\bm{x}_r}) \, d(\bm{\Psi}, \bm{\Psi}_r).
\end{flalign} 
Then according to the model,
$\Pr(\bm{\Psi}_r,\bm{\Psi}|\mathbf{X},\mathbf{Y},{\bm{x}_r})$ is the posterior distribution given $\bm{x}_r, \mathbf{X}$, and $\mathbf{Y}$ (which can be calculated from Eq. 
\eqref{eq:postint2}). The term $\Pr (y_r=m|\bm{\Psi},\bm{\Psi}_r)$ can be calculated given that $y_r$ follows a Categorical distribution with parameter vector $\bm{h}_i$, which can be directly calculated from 
$\bm{\Psi}_r$ (i.e. $\Pr (y_r=m|\bm{\Psi},\bm{\Psi}_r)=h_i(m)$).

Now, if $(\bm{\Psi}_r^{(k)}, \bm{\Psi}^{(k)})$ is the output of the MCMC at the $k$th iteration, $k=\{1,_{\cdots},K\}$ (see Section \ref{sec:MWG}), the posterior predictive distribution can be estimated using
\begin{flalign}\label{psieqn}
& \Pr(y_r=m|{\bm{x}_r},\mathbf{X},\mathbf{Y}) \\ \nonumber &\approx \frac{1}{K} \sum \limits_k \Pr (y_r=m|\bm{\Psi}_r^{(k)},\bm{\Psi}_r{^{(k)}}) \Pr(\bm{\Psi}_r^{(k)},\bm{\Psi}^{(k)}|\mathbf{X},\mathbf{Y},{\bm{x}_r}).
\end{flalign}

For Type III predictions, we simply input $(\mathbf{X},\mathbf{Y},{\bm{x}_r},y_r)$ in to the model in the learning process and then use the posterior predictive distribution in Eq. \eqref{psieqn} to carry out inference for the parameters of interest. 

\vspace{-.1in}
\appendices 
\section{Simplification of the Posterior Distribution} \label{app:details}

After performing the integrations discussed above, we will be left only with $\bm{q}$, $\bm{z}$, and $\bm{w}$.
For the first term, we have 
\begin{flalign}\label{eq:wq}
\int \Pr(\bm{w}|\bm{\tilde{q}},\bm{\vartheta}) \Pr(\bm{\tilde{q}}|\bm{q},\bm{\vartheta}) \, d\bm{\tilde{q}}	&= \int_0^1 \prod_b \prod_j \Pr(w_{:bj}|\tilde{q}_{bj}) \Pr(\tilde{q}_{bj}|q_j) \, d{\tilde{q}_{bj}} \\ \nonumber &=
\prod_b \prod_j \int_0^1 (\Pr(w_{:bj}|\tilde{q}_{bj})) \Pr(\tilde{q}_{bj}|q_j) \, d{\tilde{q}_{bj}} \\ \nonumber &= \Pr(\bm{w}|\bm{q},\bm{\vartheta})
.
\end{flalign}
As stated earlier, $\Pr(w_{ibj}|\tilde{q}_{bj})$ has a Bernoulli distribution with parameter $\tilde{q}_{bj}$ and $\Pr(\tilde{q}_{bj}|q_j)$ is a Beta distribution with parameters ($\sigma_2 \,q_j/(1-q_j),\sigma_2$). Since the posterior predictive distribution of a Binomial random variable with a Beta distribution prior on the success probability is a Beta-Binomial distribution, the inner integral in Eq. \eqref{eq:wq} becomes

\vspace{-.1in}
\begin{flalign}\label{eq:simpwapp}
	 \prod_i \Pr(w_{ibj}|q_{j}) & = \int_0^1 (\prod_i \Pr(w_{ibj}|\tilde{q}_{bj})) \Pr(\tilde{q}_{bj}|q_j) \, d{\tilde{q}_{bj}}
\\ \nonumber &=
 \frac {\mathbf{B}(q'_j+\sum \limits_i \mathbf{1}\{w_{ibj}=1\},\sigma_2+N - \sum \limits_i \mathbf{1}\{w_{ibj}=1\})} 
{\mathbf{B}(q'_j,\sigma_2)},
\end{flalign}
where $q'_j=\sigma_2 q_j/(1-q_j)$ and $\mathbf{B}(a,b)$ is the Beta function with parameters a and $b$. We can now use Eq. \eqref{eq:simpwapp} in Eq. \eqref{eq:wq} to find $\Pr(\bm{w}|\bm{q},\bm{\vartheta})$. At this point, the full posterior calculation no longer involves $\tilde{q}$. Thus, 
$$ \Pr(\bm{w}|\bm{q},\bm{\vartheta})= \prod_b \prod_j \big(\Pr(w_{:bj}|q_{j})\big).
 $$

By the conjugate property of the Beta distribution, we know that 
\begin{equation}\label{eq:qw}	
\Pr(\tilde{q}_{bj}|w_{:bj},q_j) \sim \text{Beta} (q'_j+\sum \limits_i \mathbf{1}_{\{w_{ibj}=1\}},N - \sum \limits_i \mathbf{1}_{\{w_{ibj}=1\}} + \sigma_2).
\end{equation}
From the mean of the Beta distribution given in Eq. \eqref{eq:qw}, we can conclude that given $q_j$ and vector $[w_{1bj},\cdots,w_{Nbj}]$, 
\begin{equation*}
\mathbb{E}({\tilde{q}}_{bj})=\frac{q'_j+ \sum \limits_i \mathbf{1}_{\{w_{ibj}=1\} }}{q'_j+\sigma_2+N}.	
\end{equation*}
 For the second term of Eq. \eqref{eq:postint2}, we have
\begin{flalign} \label{eq:xphikol}
\int \Pr(\mathbf{X}|\bm{\phi}) \Pr(\bm{\phi}|\bm{w},\bm{z}) d\bm{\phi} &= \int \prod_i \prod_j \Pr(x_{ij}|\bm{\phi}_{ij}) \Pr(\bm{\phi}_{ij}|w_{i:j},z_{i:}) d\bm{\phi}_{ij} \\ \nonumber
&=\prod_i \prod_j \int \Pr({x}_{ij}|\bm{\phi}_{ij}) \Pr(\bm{\phi}_{ij}|w_{i:j},z_{i:}) d\bm{\phi}_{ij} \\ \nonumber &=\Pr(\mathbf{X}|\bm{z},\bm{w},\bm{\vartheta}).
\end{flalign}
We know that $\Pr(\bm{\phi}_{ij}|w_{i:j},z_{i:})$ is a Dirichlet distribution, $\text{\text{Dir}}(\bm{g}_{ij})$, and $\Pr(x_{ij}|\bm{\phi}_{ij})$ is a Categorical distribution, $\text{Cat}(\bm{\phi}_{ij})$. Since the posterior predictive distribution of a Categorical distribution with a Dirichlet distribution prior on the probability vector is a Dirichlet-Categorical distribution, thus the inner integral of Eq. \eqref{eq:xphikol} becomes 

\begin{flalign}\label{eq:xphiapp}
 \Pr(x_{ij}|w_{i:j},z_{i:}) = \int \Pr(x_{ij}|\bm{\phi}_{ij}) \Pr(\bm{\phi}_{ij}|w_{i:j},z_{i:}) d\bm{\phi}_{ij} = \frac{g_{ij(x_{ij})}+1}{\sum \limits_v g_{ij}(v)+1}.
\end{flalign} 
We can now use Eq. \eqref{eq:xphiapp} in Eq. \eqref{eq:xphikol} to find $\Pr(\mathbf{X}|\bm{z},\bm{w},\bm{\vartheta})$. At this point, the calculation of the full posterior no longer involves $\bm{\phi}$. Since the Dirichlet distribution is the conjugate prior of the Categorical distribution, we have 
\begin{equation}\label{eq:simphix}
\Pr(\bm{\phi}_{ij}|{x}_{ij},w_{i:j},z_{i:}) \sim \text{Dir} (\mathbf{g}_{ij}(:)+\bm{\delta}_1), 
\end{equation}
where $\delta_1(v)=\mathbf{1}_ {\{x_{ij}=v\}}$. We can conclude from Eq. \eqref{eq:simphix} that given $x_{ij}$, vector $[w_{i1j},_{\cdots},w_{iSj}]$, and vector $[z_{i1},_{\cdots},z_{iS}]$, we can estimate $\mathbb{E}({\phi}_{ij}(v))$ as follows: 
 \[ \mathbb{E}({\phi}_{ij}(v))=\frac{g_{ij}(x_{ij})+ \mathbf{1}_{\{x_{ij}=v\}} }{\sum \limits_v g_{ij}(v)+1}.\]
For the third term in Eq. \eqref{eq:postint2}, we have
\begin{flalign}\label{eq:yphi}
\int \Pr(\mathbf{Y}|\bm{\theta}) \Pr(\bm{\theta}|\bm{z}) d\bm{\theta} 	&= \int \prod_i \Pr({y}_i|\bm{\theta}_i) \Pr(\bm{\theta}_i|z_{i:}) d\bm{\theta} \\ \nonumber 
&= \prod_i \int \Pr({y}_i|\bm{\theta}_i) \Pr(\bm{\theta}_i|z_{i:}) d\bm{\theta}_i \\ &= \nonumber \Pr(\mathbf{Y}|\bm{z},\vartheta). 
\end{flalign}
We know that $\Pr(\bm{\theta}_i|\bm{z}_{i})$ follows a Dirichlet distribution, $\text{Dir}(\bm{h}_{i})$, and $\Pr({y}_i|\bm{\theta}_i)$ follows a Categorical distribution, 
$\text{Cat}(\bm{\theta_{i}})$. 
Since the posterior predictive distribution of a Categorical distribution with a Dirichlet distribution prior on the probability vector is a Dirichlet-Categorical distribution, the inner integral in Eq. \eqref{eq:yphi} becomes
\begin{flalign}\label{eq:simyphiapp}
\Pr({y}_i|z_{i:}) = \int \Pr({y}_i|\bm{\theta}_i) \Pr(\bm{\theta}_i|z_{i:}) d\bm{\theta}_i = \frac{h_{i}(y_{i})+1}{\sum \limits_m h_{i}(m)+1}.
\end{flalign}
 We can now use Eq. \eqref{eq:simyphiapp} in Eq. \eqref{eq:yphi} to find $\Pr(\mathbf{X}|\bm{z},\bm{w},\bm{\vartheta})$. At this point, the calculation of the full posterior no longer involves $\bm{\theta}$.

Since the Dirichlet distribution is the conjugate prior of Categorical distribution, we have 
\begin{equation}\label{eq:simphiy}
\Pr(\bm{\theta}_i|{y}_{i},z_{i:}) \sim \text{Dir} (\mathbf{h}_{i}(:)+\bm{\delta}_2), 
\end{equation}
where $\delta_2(m)=\mathbf{1}_ {\{y_{i}=v\}}, \forall m \in \{{1,_{\cdots},M}\}$.
 We can conclude from Eq. \eqref{eq:simphiy} that given $y_{i}$ and vector $[z_{i1},_{\cdots},z_{iS}]$, we can estimate $\mathbb{E}({\theta}_{i}(m))$ as
\[\mathbb{E}({\theta}_{i}(m))=\frac{h_{i}(y_{i})+ \mathbf{1}\{m=y_{i}\} }{\sum \limits_{m} h_{i}(m)+1}. \] 
Now that the latent factors $\bm{\phi},\bm{\theta}, \text{and}, \bm{\tilde{q}}$ are integrated out, we can use a sampling process for estimating the rest of the parameters.

\vspace{-.1in}
\section{Posterior for Model II}\label{app:ModelIIPosterior}

The posterior becomes
\begin{multline}
	\Pr(\bm{q},\bm{\tilde{q}},\bm{z},\bm{w},\bm{\phi},\bm{\theta}|\mathbf{X},\mathbf{Y},\bm{\vartheta}) \propto 
\Pr(\mathbf{X},\mathbf{Y}|\bm{q},\bm{\tilde{q}},\bm{z},\bm{w},\bm{\phi},\bm{\theta},\bm{\vartheta}) \times \Pr(\bm{q},\bm{\tilde{q}},\bm{z},\bm{w},\bm{\phi},\bm{\theta}|\bm{\vartheta}) = \\
\Pr(\mathbf{Y}|\bm{\theta},\bm{w}) \times \Pr(\mathbf{X}|\bm{\phi}) \times \Pr(\bm{\phi}|\bm{w},\bm{z}) \times
\Pr(\bm{\theta}|\bm{w},\bm{z}) \times \Pr(\bm{w}|\bm{\tilde{q}}) \times \Pr(\bm{\tilde{q}}|\bm{q}) \times \Pr(\bm{q}) \times \Pr(\bm{z}).
\end{multline}
The conditional distribution for $\bm{q}$ remains the same as the model described in Section \ref{sec:fullhier}, however for $\bm{z}$ and $\bm{w}$, the conditional distributions become 
\begin{flalign}
\Pr(z_{ib}=k|z_{ib}^-) & \propto \Pr(z_{ib}=k) \times \prod_j \Pr(x_{ij}|w_{ibj},z_{ib}) \times \Pr({y}_i|w_{i::},z_{ib}), \\ \nonumber &=
\alpha^k (1-\alpha)^{(1-k)} \prod \limits_j \frac{g^{new}_{ij}(x_{ij})+1}{\sum \limits_v g^{new}_{ij}(v)+1}
 \frac{h^{new}_{i}(y_{i})+1}{\sum \limits_m h^{new}_{i}(m)+1},
 \end{flalign}
\begin{flalign}
& \Pr(w_{ibj}=k|w_{ibj}^-) \\ \nonumber
 &\propto \Pr(w_{ibj}=k|q_j) \times \Pr({x}_{ij}|w_{i:j},z_{i:},w_{ibj}=k) \times \Pr({y}_{i}|w_{i:j},z_{i:},w_{ibj}=k) \\ \nonumber
	& = \frac{\textbf{B}(q'_j+\sum \limits_i \mathbf{1}\{w_{ibj}=1\},\sigma_2+N - \sum \limits_i \mathbf{1}\{w_{ibj}=1\}} {\textbf{B}(q'_j,\sigma_2)}) \times \frac{g^{new}_{ij}(k)+1 }{\sum \limits_v g^{new}_{ij}(v)+1} \times \frac{h^{new}_{i}(y_{i})+1}{\sum \limits_m h^{new}_{i}(m)+1} .
\end{flalign}

\vspace{-.15in}
\section{Model III} \label{app:modelIII}

\begin{figure}[ht]
\centering
 \includegraphics[scale=.4, trim=1cm 0cm 1.cm .5cm]{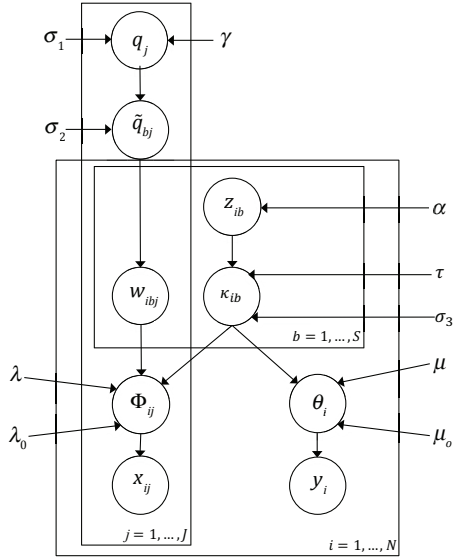}
 \caption{Directed acyclic graph for Model III. Circles indicate stochastic nodes; squares indicate fixed nodes (hyperparameters); directed edges (arrows) indicate parent-child relations.} \vspace{-.15in}
 \label{fig:ext2}
\end{figure}

In this model, the relationship between parent $b$ and individual $i$ can have a degree/weight between 0 and 1 following a Beta distribution. This is different from both Models I and II where $z_{ib}$ is a binary variable. Based on this relationship, if $i$ and $b$ are neighbors (i.e., $z_{ib}$=1), then $\kappa_{ib} \in (0,1)$ reflects the degree of connection. On the other hand, if $i$ and $b$ are not neighbors, then $\kappa_{ib}=0$. It is assumed that any nonzero $\kappa_{ib}$ follows a Beta distribution with hyperparameters $(\tau,\sigma_3)$. Note that the influences of neighbors on the generation of feature values and health labels are the same as Eqs. \eqref{eq:gnew}, and \eqref{eq:hnew}, respectively. All other parameters of the model are the same as in Model I. The directed acyclic graph of this model is given in Fig. \ref{fig:ext2}. 
Based on this model, the model hyperparameters are denoted as $\bm{\vartheta}=\{\alpha, \gamma, \sigma_1, \sigma_2, \lambda_0, \lambda, \mu_0, \mu ,a , b\}$. Given a set of known $(\mathbf{X},\mathbf{Y})$ and hyperparameters $\bm{\vartheta}$, the full posterior can be computed as:
\begin{multline}
	\Pr(\bm{q},\bm{\tilde{q}},\bm{\kappa},\bm{z},\bm{w},\bm{\phi},\bm{\theta}|\mathbf{X},\mathbf{Y},\bm{\vartheta}) \propto 
\Pr(\mathbf{X},\mathbf{Y}|\bm{q},\bm{\tilde{q}},\bm{\kappa},\bm{z},\bm{w},\bm{\phi},\bm{\theta},\bm{\vartheta}) \times \Pr(\bm{q},\bm{\tilde{q}},\bm{\kappa}, \bm{z},\bm{w},\bm{\phi},\bm{\theta}|\bm{\vartheta}) = \\
\Pr(\mathbf{Y}|\bm{\theta}) \times \Pr(\mathbf{X}|\bm{\phi}) \times \Pr(\bm{\phi}|\bm{w},\bm{\kappa}) \times
\Pr(\bm{\theta}|\bm{\kappa}) \times \Pr(\bm{w}|\bm{\tilde{q}}) \times \Pr(\bm{\kappa}|\bm{z}) \times \Pr(\bm{\tilde{q}}|\bm{q}) \times \Pr(\bm{q}) \times \Pr(\bm{z}).
\end{multline} \vspace{-.05in}
By repeating the same steps as we did in Section \ref{sec:fullhier} to omit $(\tilde{q}, \bm{\theta}, \bm{\phi})$, we get 
\begin{flalign}	\label{eq:postint}
&\int \int \int \Pr(\bm{q},\bm{\tilde{q}},\bm{\kappa},\bm{z},\bm{w},\bm{\phi},\bm{\theta}|\mathbf{X},\mathbf{Y},\bm{\vartheta}) \, d\bm{\tilde{q}}\, d\bm{\theta} \, d\bm{\phi} & \\ \nonumber 
 & =\Pr(\bm{w}|\bm{q}) \Pr(\mathbf{X}|\bm{\kappa},\bm{w}) \Pr(\mathbf{Y}|\bm{\kappa}) \Pr(\bm{\kappa}|\bm{z}) \Pr(\bm{z}) \Pr(\bm{q}). 
\end{flalign}
In the sampling process, $\bm{q}$ and $\bm{w}$ are generated the same as the original model, except that $\bm{g}$ and $\bm{h}$ are calculated according to Eq. \eqref{eq:gnew} and Eq. \eqref{eq:hnew}. The conditional distribution of $\kappa_{ib}$ and $z_{ib}$ are as follows: 
\begin{flalign}
\Pr(\kappa_{ib}|\kappa_{ib}^-) \propto \Pr(\kappa_{ib}|z_{ib}) \times \prod_j \Pr({x}_{ij}|w_{i:j},\kappa_{i:}) \times \Pr({y}_i|\kappa_{i:}). 
\end{flalign}
Given the relationship between $\kappa$ and $z$, we get 

\begin{equation}
\Pr(\kappa_{ib}=\kappa|\kappa_{ib}^-) \propto \left\{ \begin{array}{ll}
 \text{Beta}(k;\tau,\sigma_3) \times \prod \limits_j \frac{g^{new}_{ij}(x_{ij})+1 }{\sum \limits_v g^{new}_{ij}(v)+1}
 \frac{h^{new}_{i}(y_{i})+1}{\sum \limits_m h^{new}_{i}(m)+1}
 & \mbox{if $z_{ib}=1$ and $\kappa \in [0,1]$};\\
 1 & \mbox{if $z_{ib}=0$ and $\kappa=0$} \\
 0 & \mbox{if $z_{ib}=0$ and $\kappa\neq0$}
.\end{array} \right.
\end{equation}
For the elements of $z$, we have
\begin{flalign}
\Pr(z_{ib}=k|z_{ib}^-) \propto \Pr(z_{ib}=k) \times \Pr(\kappa_{ij}|z_{ib}) ,
\end{flalign}
which gives
\begin{equation}
\Pr(z_{ib}=k|z_{ib}^-) \propto \left\{ \begin{array}{ll}
 0 
 & \mbox{if $\kappa_{ib}\neq0,k=0$},\\
1 
 & \mbox{if $\kappa_{ib}\neq0,k=1$}
\\
(1-\alpha) 
 & \mbox{if $\kappa_{ib}=0,k=0$}\\
\alpha 
 & \mbox{if $\kappa_{ib}=0,k=1$}\end{array}. \right.
\end{equation}

\section{More Comparison with Decision Tree}\label{sec:appD}

\begin{figure}[ht]
\centering
 \includegraphics[scale=.44,trim=3cm .6cm 1cm 0.5cm]{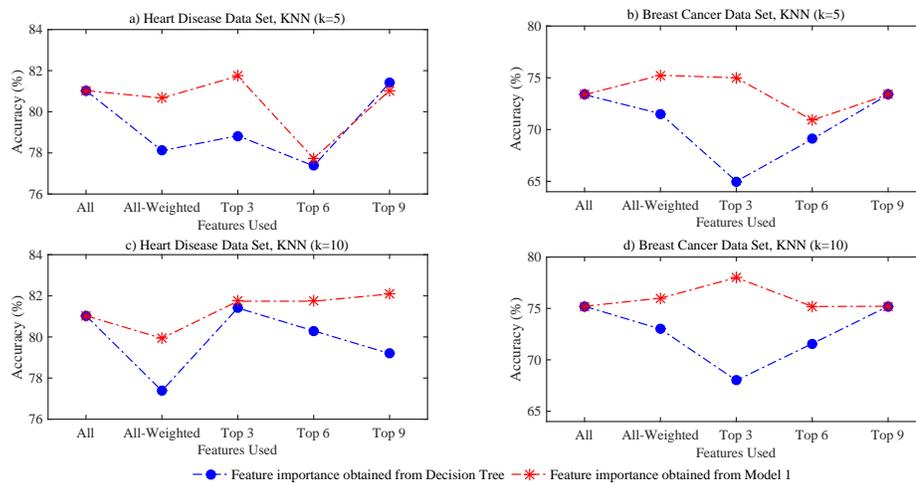} \vspace{-.1in}
 \caption{Comparison between the feature importance and ranking provided by our model and decision tree}
 \label{fig:comd32}
\end{figure}

To further evaluate the results given in Fig. \ref{fig:comd31}, we evaluate our models under the following 5 cases, (i) all features with equal weights, (ii) features with weights estimated from each model, (iii) the top three features, (iv) the top six features, and (v) the top nine features. To have a fuller comparison, a different method, here KNN, is used for classification. Results (in terms of accuracy) given in Fig. \ref{fig:comd32} verifies that weights and feature ranking from our model are comparable to the ones from the decision tree in both data sets.

\vspace{-.4in}
 \bibliographystyle{unsrtnat}
\bibliography{IEEEabrv,reference}

\end{document}